%% file: tacl2021.tex
\documentclass[11pt,a4paper]{article}
\usepackage{times,latexsym}
\usepackage{url}
\usepackage[T1]{fontenc}
\usepackage{comment}

\usepackage{xspace}
\usepackage{enumitem}

\usepackage{pgfplotstable} 
\usepackage{microtype}
\usepackage{hyperref}
\usepackage{url}
\usepackage{booktabs}
\usepackage{etoolbox}
\usepackage{amssymb}
\usepackage{float} 
\usepackage{graphicx}
\usepackage{longtable} 
\usepackage{pdflscape} 
\usepackage{adjustbox} 
\usepackage{multirow}

\usepackage{lineno}
\usepackage{xcolor,colortbl}
\usepackage[dvipsnames]{xcolor}

\usepackage[table]{xcolor}
\usepackage{makecell}
\usepackage{amsmath} 
\usepackage{soul}

\usepackage{siunitx}
\usepackage{caption}

\usepackage{lipsum}  

\usepackage{ifthen}
\usepackage{calc}
\usepackage{xparse}

\usepackage{array}
\usepackage{makecell}  
\usepackage{tabularx} 
\usepackage{subcaption}

\usepackage{listings}
\usepackage{spverbatim}
\usepackage{fancyvrb}
\usepackage{fvextra}
\usepackage{wrapfig}
\usepackage{array}
\usepackage[table]{xcolor}

\usepackage[textsize=scriptsize,backgroundcolor=green]{todonotes}
\definecolor{capri}{rgb}{0.0, 0.75, 1.0}
\newcommand\ab[1]{\todo[color=BurntOrange]{{\bf AB}: #1}}
\newcommand\eh[1]{\todo[color=WildStrawberry]{{\bf EH}: #1}}
\newcommand\ml[1]{\todo[color=capri]{{\bf ML}: #1}}
\newcommand\oa[1]{\todo[color=green]{{\bf OA}: #1}}
\renewcommand\ab[1]{}
\renewcommand\eh[1]{}
\renewcommand\ml[1]{}
\renewcommand\oa[1]{}

\definecolor{darkblue}{rgb}{0, 0, 0.5}
\hypersetup{colorlinks=true, citecolor=darkblue, linkcolor=darkblue, urlcolor=darkblue}
\usepackage[acceptedWithA]{tacl2021v1}

\usepackage{xspace,mfirstuc,tabulary}

\newif\iftaclinstructions
\taclinstructionsfalse 
\iftaclinstructions
\renewcommand{\confidential}{}
\renewcommand{\anonsubtext}{(No author info supplied here, for consistency with
TACL-submission anonymization requirements)}
\newcommand{\instr}
\fi

\iftaclpubformat 

\else

\fi

\title{PIE: Performance Interval Estimation for Free-Form Generation Tasks}

\author{%
  Chi-Yang Hsu$^{*\,\dagger}$ \quad
  Alexander Braylan$^{*\,\dagger}$ \quad
  Yiheng Su$^{\dagger}$ \quad
  Matthew Lease$^{\dagger}$ \quad
  Omar Alonso $^{**\,\ddagger}$\\
  $^{\dagger}$The University of Texas at Austin, Austin, TX, USA \\
  $^{\ddagger}$Amazon, Palo Alto, CA, USA \\
  \texttt{\{ch52669, sam.su, ml\}@utexas.edu}, \quad \texttt{braylan@cs.utexas.edu}, \\
  \texttt{omralon@amazon.com}%
}

\microtypesetup{nopatch=footnote}

\date{}

\pgfplotsset{compat=1.18}

\begin{document}
\maketitle

\begingroup
\renewcommand\thefootnote{*}
\footnotetext{Both authors contributed equally.}

\renewcommand\thefootnote{**}
\footnotetext{Work does not relate to the author's position at Amazon.}

\endgroup

\begin{abstract}
Confidence estimation infers 
a probability for whether each model output 
is correct or not. While predicting such binary correctness is sensible for tasks with exact answers, free-form generation tasks are often more nuanced, with output quality being both fine-grained and multi-faceted. 
We thus propose {\em Performance Interval Estimation} (PIE) to predict both: 1) point estimates for any arbitrary set of continuous-valued evaluation metrics; and 2) calibrated uncertainty intervals around these point estimates. 
%
We then compare two approaches: LLM-as-judge vs.\ classic regression with confidence estimation features. 
Evaluation over 11 datasets spans summarization, translation, code generation, function-calling, and question answering. Regression is seen to achieve both: i) lower error point estimates of metric scores; and ii) well-calibrated uncertainty intervals. To support reproduction and follow-on work, we  share our data and 
code.\footnote{Data and code will be shared upon paper acceptance.}

\end{abstract}


\section{Introduction}
\label{sec:intro}

Large language models (LLMs) are fallible: the quality of their outputs varies substantially across inputs, tasks, models, and evaluation metrics \cite{schellaert2025analysing}. As LLMs are increasingly applied to real-world tasks, the ability to predict the quality of each individual output 
is important to guide appropriate trust in model outputs for both human users and downstream agents  \cite{steyvers2025large, dhuliawala-etal-2023-diachronic, bhatt2021uncertainty,mielke-etal-2022-reducing, yoffe2024debunc}. 

{\bf Confidence estimation (CE)} canonically estimates the probability that a model output is correct or not. 
Predicting such binary
correctness is sensible for tasks with exact answers, but modern free-form generation
tasks are often more nuanced (e.g., summarization, translation, code generation, etc.). 
With such  tasks, output quality often
varies both 1) continuously; and 2) along multiple aspects (aka dimensions or properties), depending on the task. Example aspects might include properties such as 
fluency, relevancy \citep{liu-etal-2023-g}, factual consistency \citep{min-etal-2023-factscore}, etc. 
Predicting only unidimensional binary correctness can thus fail to capture more nuanced differences and other important dimensions of output quality. 

Why represent output quality as continuous? First and foremost, fine-grained evaluation metrics typically yield continuous scores; we want to be able to predict any arbitrary metric. Second, while subjective human quality ratings may be labeled ordinally, more objective quality measures are naturally continuous (e.g.,  execution time of generated code or time required to further revise model outputs). Framing the quality prediction task as regression allows us to be agnostic as to both: 1) whether the prediction target is ordinal or continuous; and 2) whether we are predicting an observed value (e.g., human rating or program execution time) or a derived evaluation metric. 

%

This regression framing also has important implications for {\bf uncertainty quantification (UQ)}. When quality is binary, we can formulate uncertainty as the probability of an output being correct or not. However, when output quality is continuous, UQ must also shift to estimating confidence intervals around point estimates of output quality. 

Such UQ is particularly consequential in high-stakes settings where decisions are made directly from model outputs, necessitating risk-aware judgments. For example, consider software development workflows where LLM-generated code is increasingly executed or deployed with limited human oversight
\citep{cotroneo2025human,navneet2025rethinkingautonomypreventingfailures}. High confidence is required that generated code will reliably execute, pass unit tests, and be efficient and robust \citep{tong-zhang-2024-codejudge}\footnote{Predicting generated code quality pre-execution avoids both potential security risks and any time/space requirements.}.
%
%
To enable risk-informed decision-making, a point estimate of output quality alone is insufficient; a confidence interval helps lower-bound quality to better assess the level of risk.  

For example, imagine generated code has predicted functional correctness of 0.89 with a narrow 95\% interval (e.g., $\pm 0.09$), meaning 95\% confidence that output quality is $\ge$ 80\%. Assuming point estimates and confidence intervals are well-calibrated (i.e., trustworthy), this  may be considered sufficiently low risk to use the output without human review. Alternatively, if the same predicted correctness of 0.89 were instead accompanied by a wide interval (e.g., $\pm 0.28$), this would indicate substantial uncertainty and a more serious worst-case scenario: i.e., 95\% confidence that output quality is only $\ge$ 0.61. In practice, such uncertainty can translate into failures that are costly to detect and resolve,
motivating additional safeguards, such as requiring human review and revision. \textbf{Table~\ref{tab:example_data}} illustrates such an example in more detail. Prediction intervals are thus essential for guiding risk-aware decision making \citep{hullman2025conformal}. Intervals must also be well-calibrated, since miscalibrated intervals can do harm in leading users astray. 


In sum, this work identifies the need for: 1) supporting arbitrary, task-specific measures of output quality for nuanced, free-form generation tasks; 2) continuous, multidimensional point estimates of output quality; and 3) confidence intervals around point estimates. To address these needs, we propose 
\textbf{P}erformance \textbf{I}nterval \textbf{E}stimation (\textbf{PIE}). The goal of PIE is to enable calibrated trust in model outputs for such tasks via a combination of predicting: 
1) point estimates of continuous, task-specific quality measures; and 2) calibrated confidence intervals around each point estimate. 

By jointly modeling quality and uncertainty, PIE supports 
practical deployment decisions such as abstention, asking  clarification questions, or fallback to alternative models in low-confidence cases \citep{10.1145/3331184.3331265, wen-etal-2025-know,  chuang2025learning}. High-quality outputs may be accepted directly, while low-quality ones can be revised, regenerated, or discarded. Predicted quality could further support prompt or model selection \cite{zhou2022least}, candidate ranking \cite{chen-etal-2023-adaptation}, and routing inputs to models  expected to perform best \cite{lee-etal-2024-orchestrallm, vsakota2024fly, lu2023routing, shnitzer2023large, dinghybrid}.

For reproducible benchmarking on PIE's prediction objective, we identify a set of 11 datasets spanning summarization, machine translation, function-calling, code generation, and question answering. Since different evaluation metrics capture different aspects of output quality, we include multiple metrics per task. To benchmark over outputs from different generative models, we include outputs from Llama 3.2-11B and Gemini 2.0-flash-lite.

We then evaluate two approaches: reference-free LLM-as-judge vs.\
classic regression using confidence estimation features. For the former, we prompt an LLM to directly predict the target evaluation metric score and a corresponding prediction interval for each generated output. Our evaluation includes use of in-context learning with a small set of labeled examples. The classic regression model predicts metric scores and intervals from confidence estimation signals, including verbalized confidence, consistency-based measures, and other uncertainty  methods, by training a regression model to map these features to continuous metric targets.

Results across the 11 datasets, 15 metrics, and 2 LLMs show that the regression-based approach performs best in estimating both point estimates and confidence intervals for metric scores. 
Because free-form gold references can be expensive to collect, sample-efficiency is also important. Results show the regression-based approach converges after only 16 training instances across all 11 datasets.   

Overall, 
we conceptually integrate three theoretically related but methodologically distinct lines of prior work:  confidence estimation, quality estimation, and LLM-as-a-judge (Section \ref{sec:relation}). In reviewing each line's core assumptions, methodological choices, and limitations, we show none provides a task-agnostic, general solution for predicting uncertainty in free-form generation. We then introduce a unified task formulation, PIE, to predict continuous evaluation metrics together with calibrated uncertainty intervals (Section \ref{sec:predict_problem}). Since PIE is agnostic regarding evaluation based on directly observed values or derived performance metrics, it is equally suited to predicting both reference-based quality metrics (e.g., based on human ratings) as well as more objective metrics like code run-time, memory use, or tool call alignment.  

To enable reproducible benchmarking, we collate and share a multi-task, multi-metric benchmark dataset  (Section \ref{sec:predict}). Experiments 
(Section~\ref{sec:tasks}) and findings (Section~\ref{sec:results_findings}) shows that classic regression matches or exceeds contemporary LLM-as-judge methods, consistently achieving more accurate and better-calibrated intervals while also using as few as 16 labeled examples to do so (Section \ref{sec:results_findings}).

\begin{table}[!t]
\centering
\scriptsize
\setlength{\tabcolsep}{3pt}
\renewcommand{\arraystretch}{1.15}
\begin{tabular}{|p{0.24\columnwidth}|p{0.72\columnwidth}|}
\hline
\textbf{Input} &
Scale the input field to the range [0, 1] and display it as a DataFrame.\\
\hline
\textbf{LLM output} &
{\ttfamily
scaler = MinMaxScaler( feature\_range=(0, 1))
\newline
\ \ \ \ scaled\_values = scaler.fit\_transform([[x] for x in l])
\newline
\ \ \ \ df = pd.DataFrame(scaled\_values, columns=['Scaled Values'])
\newline
\ \ \ \ return df
} \\
\hline
\textbf{Target text} &
{\ttfamily
scaler = MinMaxScaler()
\newline
l\_scaled = scaler.fit\_transform( l.reshape(-1, 1))
\newline
df = pd.DataFrame(l\_scaled, columns=['Scaled Values'])
\newline
return df
} \\
\hline
\textbf{Prediction} & $0.8918 \pm 0.2824$  at 95\% probability \\
\hline
\textbf{Gold score} & $0.7500$ (functional correctness) \\
\hline
\end{tabular}
\caption{
Illustrative PIE example on BigCodeBench. The CE-Reg predictor estimates functional correctness and a 95\% prediction interval. 
This interval covers the gold score despite an optimistic point estimate, mitigating overconfident assessment of code that could fail for array-shaped inputs.
}
\label{tab:example_data}
\end{table}

\begin{table*}[h!]
\centering
\caption{Comparison of output quality estimation paradigms, illustrating their respective strengths and limitations in supporting continuous, multi-dimensional evaluation and uncertainty for free-form generation.}
\label{tab:comparison}
\begin{tabular}{|p{0.15\linewidth}|p{0.17\linewidth}|p{0.18\linewidth}|p{0.18\linewidth}|p{0.19\linewidth}|}
\hline
& \makecell[l]{\textbf{Confidence}\\\textbf{Estimation}}
& \makecell[l]{\textbf{Quality}\\\textbf{Estimation}}
& \makecell[l]{\textbf{Reference-free}\\\textbf{LLM-as-a-Judge}}
& \makecell[l]{\textbf{Performance }\\\textbf{Interval Estimation}}
\\
\hline
\textbf{Used mostly in}
& Limited (QA)
& Limited (MT)
& Various tasks
& Various tasks \\
\hline
\makecell[l]{\textbf{Continuous}\\\textbf{targets}}
& No
& Yes
& Yes
& Yes \\
\hline
\makecell[l]{\textbf{Multiple}\\\textbf{dimensions}}
& No
& \makecell[l]{{Limited}\\{(error types)}}
& Yes
& Yes \\
\hline
\makecell[l]{\textbf{Explicit}\\\textbf{uncertainty}}
& Yes
& Limited
& Limited
& Yes \\
\hline
\end{tabular}
\end{table*}

\section{Related Work}
\label{sec:relation} 

In this section, we review and compare key related work in confidence estimation (Section~\ref{sec:uncertainty}), quality estimation (Section~\ref{sec:mt}), and (reference-free) LLM-as-judge (Section~\ref{sec:LLMaaJ}). 
%
%
%
Each of these varies in how uncertainty and metric prediction are considered. 
To relate these approaches and their relative capabilities, we compare them via several criteria in particular regard to evaluating the quality of free-form generations, as summarized in \textbf{Table~\ref{tab:comparison}}. 




\subsection{Confidence Estimation (CE)} \label{sec:uncertainty}

Confidence estimation canonically models the reliability of a system output as the probability of an output being correct or not, producing a single scalar confidence value tied to binary correctness \citep{Blatz2004-rp, guo2017calibration, steyvers2025large}. 
This view naturally fits short, discrete tasks with exact answers, such as factoid questions 
where correctness can be treated as all-or-nothing.

Two CE approaches are particularly well-known for their simplicity and applicability to black-box models: \emph{verbalized} CE \cite{xiong2024can, tian-etal-2023-just} and \emph{consistency-based} CE \citep{kuhn2023semantic, lin2024generating}. The former prompts an LLM to self-report its confidence alongside the generated output 
in a single LLM invocation. Consistency-based methods instead infer confidence from the degree of agreement across multiple independent generations. Computational overhead thus scales linearly with the number of samples.



As discussed earlier, framing output quality wrt.\ binary correctness is less suited to free-form, nuanced generation tasks in which assessment of output quality is often fine-grained (i.e., continuous), multi-dimensional, and task-dependent. 

\subsection{Quality Estimation (QE)} \label{sec:mt}
Quality estimation in machine translation (MT) predicts output translation quality at runtime \cite{Specia2009-nx,Specia2010-pk}. In comparison to PIE, it shares both important similarities and notable differences. For example, early work on CE for machine translation also assumed binary correctness \cite{Blatz2004-rp}, which QE then generalized to continuous, partial-credit evaluation \cite{Specia2009-nx,Specia2010-pk}. PIE is similarly inspired but extends this to diverse tasks, arbitrary metrics, and a strong emphasis on uncertainty and calibrated prediction intervals.

Work on QE has been largely restricted to MT, though other potential application areas have been discussed. For example,  \citet{Scarton2016-hh} suggested generalizing QE to tasks such as correcting grammar \cite{napoles2016there}, generation \cite{duvsek2019automatic}, and others \cite{specia2018quality}. In contrast, PIE is fundamentally conceived as task- and domain-agnostic. Canonically, QE also refers to predicting human quality ratings,  i.e., ``direct assessment'' judgments of translation quality. That said, some work has considered other metrics such as post-editing time \cite{specia2018quality}. In contrast, PIE is fundamentally metric-agnostic. While both QE and PIE are concerned with output ``quality'', PIE's notion of quality is broader, encompassing arbitrary task-dependent metrics of quality, such as the functional correctness of code.   

Another key difference between QE and PIE lies in their respective treatments of uncertainty quantification (UQ). While recent work has explored uncertainty within QE, principally via conformal prediction \citep{glushkova-etal-2021-uncertainty-aware, giovannotti2023evaluating, zerva-martins-2024-conformalizing}, even here, uncertainty is not treated as a core modeling objective, and the scope remains confined to translation. In contrast, UQ is central and fundamental to PIE to support risk-aware decision making across diverse tasks and settings. As Table~\ref{tab:example_data} illustrates, even a high predicted quality score can still present significant risk if the prediction interval is wide. 
%


Within QE, \citet{Zouhar2023-gf} proposed {\em metric estimation}: predicting reference-based evaluation metrics without access to the human reference. Similar to PIE, the focus is on predicting evaluation metrics, not just human judgments. However, like most QE work, its scope is translation-specific, focused on a small set of  reference-based metrics, and predicts only point estimates.  



\subsection{LLM-as-a-Judge (LLMaaJ)} 
\label{sec:LLMaaJ}
The rise of LLMs has also brought the {\em LLM-as-a-judge} paradigm. A key distinction is whether an LLMaaJ is used at runtime (online) to predict output quality vs.\ being used during evaluation (offline) to evaluate output quality. Closely related to this is the distinction between {\em reference-based} (RB) vs.\ {\em reference-free} (RF) methods. RB methods require a human (gold) reference output with which to compare the generated output and so are only suitable for use during evaluation (offline) \citep{Sheng2024-et, li2024llmsasjudgescomprehensivesurveyllmbased}. In contrast, RF methods have no such requirement, assessing output quality without comparing to any reference output \citep{liu-etal-2023-g}. As such, RF methods are more flexible and can be used both online and offline.

We view RF-LLMaaJ as closely related to CE and QE since all three seek to predict output quality at runtime. However, a key limitation of LLM approaches seen in prior studies is their limited ability to regress continuous values, often struggling to produce calibrated intervals \citep{vedula-etal-2025-quantile}, also seen in our own findings (Section~\ref{sec:results_findings}). 

%
%
%
RF-LLMaaJ shares conceptual similarities with “verbalized’’ CE methods (Section~\ref{sec:ce}) that prompt a model for its confidence in its output correctness \citep{tian-etal-2023-just, xiong2024can}. Both rely on prompting an LLM at runtime, but verbalized CE assumes a specific assessment target (confidence), whereas RF-LLMaaJ methods have been explored more broadly in evaluating model outputs. 

RF-LLMaaJ methods also share conceptual similarities with QE methods (Section~\ref{sec:mt}). As discussed, QE canonically seeks to predict human ratings (for translation quality) and only rarely to predict objective measures (e.g., post-editing time \cite{specia2018quality}) or evaluation metrics \cite{Zouhar2023-gf}). Similarly, RF-LLMaaJ work also typically seeks to predict human judgments and is evaluated by how well it aligns with them, rather than predicting reference-based evaluation metrics. As noted earlier, prediction of confidence intervals is atypical in QE work \citep{glushkova-etal-2021-uncertainty-aware, giovannotti2023evaluating, zerva-martins-2024-conformalizing}. Similarly, uncertainty quantification for continuous predictions (e.g., inferring confidence intervals) is also rare with LLMs \citep{hashemi-etal-2024-llm}.

As in CE and QE, a key premise of our work is that (offline) evaluation requires information not available (online) at runtime. Examples include human reference outputs or knowledge such as memory use by generated code once executed. The key challenge is thus online prediction of offline evaluation. However, when online RF-LLMaaJ is also the standard for offline evaluation, there is no need to predict it. We posit that despite growing popularity of evaluation via RF-LLMaaJ methods, many important evaluation settings will continue to require information not available at runtime. 

\section{PIE Task Formulation} 
\label{sec:predict_problem}



As summarized in Table~\ref{tab:comparison}, existing methods either model uncertainty over binary correctness or predict continuous quality without task-general uncertainty estimates. We therefore propose PIE, which given a model output, predicts a continuous evaluation metric for it, along with a calibrated uncertainty interval. PIE extends point-estimate prediction by bounding the range within which the true performance score is expected to fall with a specified probability (e.g., 95\%) \cite{SHRESTHA2006225, patel1989prediction, geisser1993predictive}.

\vspace{0.5em}
\noindent \textbf{Formalization.}
\label{sec:problem}
Let $\mathcal{X}$ denote the input space (user queries, upstream agent inputs) and $\mathcal{Y}$ the free-form output space (e.g.,\ text, code, etc.) for possible LLM generations, as well as gold reference outputs, if available. We assume each task $t \in \mathcal{T}$ has $K_t$ continuous evaluation metrics $\mathcal{M}_t = \{m_{t,1},\dots,m_{t,K_t}\}$, where each metric
$
m_{t,k}: \mathcal{Y} \rightarrow \mathbb{R}
$
assigns LLM output $y$ a real-valued quality score based on a measure of quality -- for example, direct human judgment or agreement with reference gold output $g$. 
At inference, only input $x$ and generation $y$ are observed; metric score $m_{t,k}(y)$ is not. 

\vspace{0.5em}
\noindent \textbf{Prediction.}
For each task-specific metric $m_{t,k}$, we seek a predictor defined by:
\vspace{-0.25em}
$$
f_{t,k}: (\mathcal{X}\!\times\!\mathcal{Y}) \rightarrow \big(\hat{s}, [\ell,u]\big)
$$
for instance $(x,y)$ that returns: (i) a point estimate $\hat{s}\in\mathbb{R}$ for metric score $m_{t,k}(y)$ and (ii) a $(1-\alpha)$ prediction interval $[\ell,u]\subseteq\mathbb{R}$ around $\hat{s}$ such that:
\vspace{-0.25em}
$$
\Pr\!\big(m_{t,k}(y) \in [\ell,u]\big) \ge 1-\alpha,
$$
for a specified confidence level (e.g., $\alpha=0.05$).

\vspace{0.5em}
\noindent \textbf{Supervision.}
For each task $t$, assume we have a dataset $\mathcal{D}$ containing $N_t$ training instances $\{(x_i, y_i, s_i)\}_{i=1}^{N_t}$ for each of $t$'s $K_t$ metrics,  
%
%
%
where $s_i = m_{t,k}(y_i)$ denotes the gold score for metric $m_{t,k}$, as assessed for generation $y_i$. This setup is both metric-agnostic (allows any metric) and model-agnostic (makes no use of model parameters or decoding internals). 

\vspace{0.5em}
\noindent \textbf{Objective:}
minimize point-estimation error and produce calibrated intervals (see Section~\ref{sec:assessing_performance_prediction}).

\section{Prediction Methods}
\vspace{-0.5em}
\label{sec:predict}

Having defined the PIE performance predictor $f_{t,k}$ (Section~\ref{sec:problem}), we now describe two such predictors. 


\vspace{0.5em}
\noindent \textbf{Confidence-based Regression (CE-Reg)}: Extract unsupervised confidence estimates as features, then learn a regression model to map features to the continuous metric score and a prediction interval. Sections~\ref{sec:ce} and~\ref{sec:ce-reg} provide further detail. 
    
\vspace{0.5em}
\noindent\textbf{Reference-free LLM-as-a-Judge (RF-LLMaaJ)}: Prompt an LLM to directly output a point estimate and prediction interval for the target metric, using labeled examples for in-context learning. Section~\ref{subsec:judge} provides further detail. 

%

\subsection{Confidence Estimation (CE)}
\label{sec:ce}


\textbf{Verbalized confidence} 
%
prompts an LLM for its confidence that an output is correct. One-step (1S) produces both output and estimated confidence in one invocation, while two-step (2S) first generates the output and then calls the LLM a second time to estimate output confidence. While 1S reduces computation, 2S provides cleaner delineation for benchmarking PIE methods on model fixed outputs. 
%
%
See Appendix~\ref{app:sampleprompts} for example prompts. 
    
\vspace{0.5em}
\noindent \textbf{Consistency-based}: 
We generate $n$ outputs and measure coherence between them via 5 methods: Degree Matrix (Deg; \citet{lin2024generating}), Eccentricity (Ecc; \citet{lin2024generating}), Sum of Eigenvalues of the Graph Laplacian (EigV; \citet{lin2024generating}), Lexical Similarity (LexSim; \citet{Fomicheva_tacl_a_00330}), and the Number of Semantic Sets (NumSet; \citet{lin2024generating}). Most use natural language inference (NLI) models to assess coherence between outputs, while LexSim is computed using ROUGE-L. 

\subsection{Confidence-Based Regression (CE-Reg)}
\label{sec:ce-reg}

Given a dataset 
$
\mathcal{D} = \{(c_i, s_i)\}_{i=1}^N,
$
of $N$ training examples, where each $c_i$ is the confidence feature (verbalized or consistency-based) for output $y_i$ and $s_i = m_{t,k}(y_i)$ is the 
gold score, we fit an auxiliary regression model to learn the conditional distribution
$
s_i \sim p(s \mid c_i, \theta),
$
for gold metric score $s_i = m_{t,k}(y_i)$.  From the fitted distribution we extract a point estimate $\hat{s}_i$ and a $(1-\alpha)$ prediction interval $[\,\ell_i, u_i\,]$ (using $\alpha=0.05$) via
{\setlength\abovedisplayskip{5pt}
 \setlength\belowdisplayskip{5pt}
 \[
   \ell_i = F^{-1}\bigl(\tfrac{\alpha}{2}\bigr),\quad
   u_i = F^{-1}\bigl(1 - \tfrac{\alpha}{2}\bigr)
 \]
}
\!where $F$ is the model’s predictive cumulative distribution function (CDF).  

\textbf{Regression Models.} We compare a variety of models: linear (Linear, Bayesian Ridge), tree ensembles (Random Forest, XGBoost), and a Beta regression for bounded metrics. Gaussian-error models provide mean $\hat{s}$ \& variance $\hat{\sigma}^2$ directly; ensembles approximate variance via bootstraps/trees. 

\subsection{Reference-free LLM-as-a-Judge}
\label{subsec:judge}

Building on prior work (Section~\ref{sec:LLMaaJ}), we also compare a reference-free LLM-as-a-Judge (RF-LLMaaJ) for continuous-score estimation using in-context learning \cite{Vacareanu2024-mk}. It is prompted with: 
1) a brief description of the task and the scoring metric; 
2) $N$ demonstration examples in form (input, output, gold metric score); 
3) instructions to produce a point score and 95\% upper/lower interval bound;
and 
4) A test pair (input, output) for which it must predict a point estimate and prediction interval for the target metric. 
%
%


\begin{figure*}[ht]
  \centering
  \includegraphics[width=\textwidth]{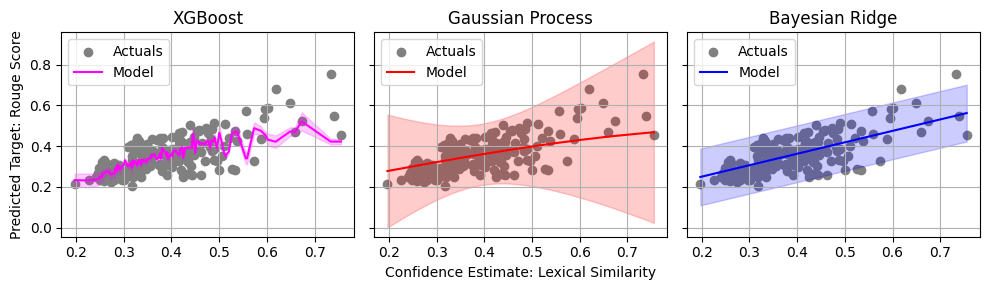}
  \vspace{-2em}
  \caption{Illustrating different regression model predictions (solid lines) and confidence intervals (color shading around the lines). In this example, a Bayesian Ridge regression model performs best as predictions are neither over-confident (XGBoost) nor under-confident (Gaussian Process).}
  \label{fig:ensemble_vs_individual}
\end{figure*}

\section{Experiments}
\label{sec:tasks}

\subsection{Experimental Setup}
\label{sec:setup}

\textbf{Datasets\,\&\,Tasks.}
Table \ref{tab:dataset_summary} details our datasets, tasks, and evaluation metrics used. Tasks span code generation, fact-checking, function-calling, machine translation (MT), question-answering (QA), text summarization, and others. We randomly sample up to 5,000 instances per dataset, splitting each 3/1/1 into train, dev, and test sets. For each question, the LLM generates three outputs. Each output is evaluated based on task-specific metrics, producing a ⟨input, LLM output, metric score⟩ tuple.

\vspace{0.5em}
\textbf{Task-Specific Evaluation Metrics.} Each task requires predicting one or more continuous metrics. Given our metric-agnostic formulation, the choice of metrics used is less important than assessing the ability to predict arbitrary metrics, including different metrics for the same generation task. 

Metrics include: BERTScore \citep{bert-score}, CodeBLEU \citep{ren2020codebleu}, and 
reference-based (RB) LLM-as-a-judge metrics, such as {LLM-Eval (accuracy and informativeness)} for QA \citep{10.1145/3726302.3729895, lin-chen-2023-llm} and BiGGen \citep{kim-etal-2025-biggen}, GEMBA\_ref (with references) \citep{kocmi-federmann-2023-large} for MT,  G-Eval (coherence, consistency, fluency, and relevancy) \citep{liu-etal-2023-g} for text summarization, FActScore \citep{min-etal-2023-factscore} for fact-checking, and CodeJudge \cite{tong-zhang-2024-codejudge} for code generation and function-calling (functional correctness and inconsistency). These LLM judge methods support more robust evaluation by considering diverse aspects of quality, such as factual correctness, informativeness, fluency, relevance, etc.

\textbf{Direct Assessment (DA) Task}. To show PIE's generality in supporting prediction of both  directly observed quantities as well as derived evaluation metrics, we include a proof-of-concept evaluation of predicting observed human quality ratings for the BiGGen-Bench-Results dataset \citep{kim-etal-2025-biggen} (Section~\ref{sec:mt}). This dataset provides human ratings for its pre-generated LLM outputs, and we use both its LLM outputs and human ratings. 
However, because the dataset only contains one output per rating, this precludes evaluation of 
consistency-based methods (see Section \ref{sec:ce}).

%

\textbf{LLMs.} Llama 3.2-11B \citep{meta2023llama32} and Gemini-2.0-Flash-Lite \citep{team2023gemini} are used to produce output generations. For PIE prediction, whichever LLM was used for generation is then also used for both predictors, CE-Reg and RF-LLMaaJ. 
We set {\tt max tokens}~$=769$, {\tt temperature}~$= 1$, and {\tt topP}~$=0.9$. To avoid any self-enhancement bias \citep{zheng2023judging} in ensuring high quality evaluation scores to predict, we use Gemini-2.5-Flash \citep{team2023gemini} for reference-based (RB) LLMaaJ evaluation metrics.

The DA task above differs in using existing LLM outputs from prior work: \citet{kim-etal-2025-biggen}'s GPT-4-0125 \citep{achiam2023gpt} generations. 

\textbf{PIE Predictors.} 
%
Consistency methods are implemented by \texttt{LM-Polygraph} \citep{fadeeva2023lm} and computed over $n\!=\!3$ outputs, with all $n$ outputs sharing the same consistency score. We implement regression models in \textsc{scikit-learn} and use XGBoost with default hyperparameters. Beta regression is implemented in-house. 
For regression model selection, we compare results on the dev set. 
Based on this (Figure \ref{fig:radar_plots}), we select Random Forest for our test set evaluation.

For RF-LLMaaJ in-context learning, we vary $N \in \{4,8,16\}$ demonstration examples. We cap at 16 demonstration examples due to high token costs (especially with free-form outputs) and potential to degrade performance \cite{levy2024same,modarressi2025nolima}. In contrast, CE-Reg is far more efficient to train at scale. 

As for relative LLM compute costs of different methods (beyond the output generation call): consistency methods require $n-1$ additional calls (to generate additional outputs for measuring output consistency), verbalized 1S requires no additional calls, verbalized 2S requires one additional call, and RF-LLMaaJ also requires one additional call. In practice, the $n$ outputs generated for consistency would be used to select the output predicted to perform best, so the added compute cost is offset by expected quality improvement in output selection.  
Appendix~\ref{app:sampleprompts} lists prompts and parsing rules. 



\input{table/ac_bc_table_crps}
\subsection{PIE Evaluation} 
\label{sec:assessing_performance_prediction}
Predictor $f_{t,k}$ (Section~\ref{sec:problem}) must return both a point estimate $\hat{s}$ and a predictive distribution (with a $(1-\alpha)$ interval). We evaluate these with 4 \textit{PIE metrics} (not to be confused with task-specific metrics).

\noindent \textbf{Continuous Ranked Probability Score (CRPS)}
\citep{gneiting2007strictly} is PIE's primary metric for evaluating both point estimates and prediction intervals. It is a proper scoring rule that jointly rewards calibration and sharpness of the full predictive CDF $F$; lower CRPS is better:

{\setlength\abovedisplayskip{5pt}
 \setlength\belowdisplayskip{5pt}
$$
\mathrm{CRPS}(F, s) = \int_{-\infty}^{\infty} \big(F(x) - \mathbf{1}\{x \ge q\}\big)^2 \, dx
$$
}

Prediction intervals for continuous scores can be visualized in Figure~\ref{fig:ensemble_vs_individual}, showing each instance's true score on the y-axis and the prediction regressor on the x-axis. Ideally, predictions are neither too narrow (overconfident) nor too wide (underconfident).
While Expected Calibration Error (ECE) is widely used for classification to compare predicted probabilities with observed frequencies via binning \citep{naeini2015obtaining}, it is traditionally defined over \emph{binary} scores, not continuous ones.
\citet{kuleshov2018accurate} propose an ECE variation for continuous scores based on quantile coverage, but unlike CRPS, such diagnostics do not reward distribution sharpness and can therefore favor overly conservative (wide) intervals \citep{gneiting2007probabilistic}.

\noindent \textbf{Other complementary metrics}
We also report \textit{RMSE} (Appendix~\ref{tab:benchmark_rmse}), providing an interpretable measure of average error magnitude in the target’s original units.  
\textit{Pearson Correlation (PC)} is also used as a diagnostic for feature usefulness (Appendix~\ref{tab:benchmark_corr}); it does not directly reflect absolute error, and correlation may be high while still incurring large systematic errors. However, as it is unscaled, it is useful for comparing between tasks and metrics and for reflecting instance-level discriminative predictability.

In addition, 
we also measure \textit{Average Coverage Error (ACE)} \citep{shrivastava2016multiobjective,xie2023overview} (Appendix~\ref{tab:benchmark_ace}), which compares empirical coverage of the nominal $(1-\alpha)$ interval to its target as $\mathrm{ACE} = \bigl|\mathrm{PICP}(p) - p\bigr|$ where
{\setlength\abovedisplayskip{5pt}
 \setlength\belowdisplayskip{5pt}
\begin{equation*}
  \begin{aligned}
    \mathrm{PICP}(p) &= \frac{1}{n}\sum_{i=1}^{n}\mathbf{1}\{\,\ell_i(p) \le s_i \le u_i(p)\,\} 
    \end{aligned}
\end{equation*}
}
and $p=1-\alpha$ (we use $p=0.95$). Lower ACE indicates better calibration, with positive $p$ reflecting under-confidence and negative $p$ reflecting over-confidence. Note that one disadvantage of ACE is that it does not penalize excessively wide intervals (e.g., under-confident intervals that always cover $s_i$ incur a maximum $0.05$ penalty when $p=0.95$) as much as narrow ones. Hence, we treat ACE as a secondary, human-readable diagnostic and rely primarily on CRPS for PIE evaluation.

\input{images/model_compare_fig}

\section{Findings} 
\label{sec:results_findings}
We now summarize 6 key empirical findings from evaluating PIE over 11 free-form generation tasks, 15 metrics, and 2 backbone LLMs. Rather than focus solely on point estimates or preference judgments, our experiments jointly assess the prediction of continuous evaluation scores and their associated uncertainty. We organize the results to study key properties of PIE, including feasibility, sample efficiency, metric-dependent behavior, and robustness across estimation techniques and models. 
\noindent \textbf{PIE is feasible}
Across datasets and target metrics, CE-based regression achieves an average CRPS of 0.10, RMSE of 0.16, and PC of 0.34 (Tables~\ref{tab:benchmark_crps}, \ref{tab:benchmark_rmse}, and~\ref{tab:benchmark_corr}), indicating that simple confidence features support useful prediction of continuous quality scores and calibrated uncertainty at inference time. This holds both for derived evaluation metrics and for human ratings (Table~\ref{tab:direct_assessment}). Performance also stabilizes with as few as 16 labeled examples, suggesting PIE can be practical when annotation budgets are small.


\noindent \textbf{PIE supports sample-efficient predictors}
As free-form gold references can be expensive to collect, it is important to understand how sample-efficient these methods would be. To answer that, we vary the size of the training set from as few as 4 instances and doubling up to the full training data size for CE-Reg. For RF-LLMaaJ, we limit up to 16 instances.
CE-Reg achieves close to our best observed accuracy using few labeled examples: Figure~\ref{fig:train_size_plot} shows that CRPS converges after just 16 training instances across all datasets.  Even when compared to RF-LLMaaJ’s maximum of 16 in-context examples (Table~\ref{tab:sample_table}), CE-Reg delivers substantially lower error.
Although performance varies slightly across datasets, the convergence point is consistent. This shows that the regression approach can be effectively trained with a small sample size, making it easy to collect training data even for out-of-domain tasks to mitigate potential domain differences.

\vspace{0.5em}
\noindent 
\textbf{PIE is more accurately addressed by CE-Reg than by RF-LLMaaJ.}
Because PIE requires predicting continuous evaluation metrics together with uncertainty at the instance level, not all evaluation paradigms are equally well matched to this objective. We therefore compare CE-Reg and RF-LLMaaJ as two representative paradigms for performance prediction under PIE.

Across datasets and evaluation metrics, CE-Reg consistently achieves substantially lower error under the primary CRPS objective than RF-LLMaaJ (Table~\ref{tab:benchmark_crps}). Averaged over task metrics, CE-Reg attains CRPS values averaging 0.11, compared to 0.14 for RF-LLMaaJ, indicating more informative predictive distributions. This advantage is also reflected in point accuracy: CE-Reg achieves RMSEs averaging 0.16, whereas RF-LLMaaJ incurs much larger errors, averaging around 0.22 (Table~\ref{tab:benchmark_rmse}). 

In addition, CE-Reg preserves stronger instance-level ordering, achieving consistently higher correlation of 0.28 across metrics than RF-LLMaaJ's 0.09 (Table~\ref{tab:benchmark_corr}). 
\textbf{Notably, for many task–metrics, RF-LLMaaJ exhibits near-zero correlation}, indicating limited ability to recover instance-level signal. Increasing in-context examples from 4-16 yields negligible improvement (Table \ref{tab:sample_table}), suggesting that additional prompting alone does not reliably address this. 

One plausible explanation is RF-LLMaaJ’s sensitivity to prompt formulation and target specification. Although both RF-LLMaaJ and verbalized confidence rely on in-context LLM predictions, verbalized confidence appears more stable across tasks, whereas RF-LLMaaJ must predict diverse task-specific metrics, which few-shot prompting may not reliably support. Taken together, these findings indicate that confidence estimates are informative signals for PIE but require supervision for effective calibration, while RF-LLMaaJ approaches remain less reliable across metrics and evaluation sources.

\noindent \textbf{PIE benefits most from graph-based consistency features.}
Among all CE-Reg, the four graph-based consistency measures (DegMat, Eccentricity, EigVal, LexSim) achieve the lowest CRPS, indicating that capturing variability across multiple generations provides a stronger predictive signal than one- or two-step verbalized confidence prompts, on all types of datasets and metrics. By contrast, the NumSemSets consistency metric performs on par with the verbalized methods, trailing the graph-based techniques. Better results from graph-based consistency over NumSemSets and verbalized confidence are more obviously reflected in the ACE and PC measures (Tables \ref{tab:benchmark_ace} and \ref{tab:benchmark_corr}).

\noindent \textbf{PIE yields consistent results across LLMs.}
Figure~\ref{fig:model_comparison_ce_vs_rf} evaluates the generality of CE-Reg and RF-LLMaaJ by comparing their performance across two backbone models. In addition to Llama~3.2~11B, we re-ran the same experimental setup on Gemini~2.0~Flash-Lite. Across both models and all evaluation criteria, we observe that CE-Reg outperforms RF-LLMaaJ, with absolute performance levels also comparable across models. This consistency suggests that the advantages of CE-Reg are not model-specific and that PIE generalizes across LLM families.

\noindent \textbf{PIE performance varies across task metrics.}
Across datasets, BERTScore and CodeBLEU achieve low CRPS (Table~\ref{tab:benchmark_crps}) and high correlation (Table~\ref{tab:benchmark_corr}), indicating that their \emph{per-instance predictive distributions} are accurate in absolute magnitude and that predicted scores are well-aligned in relative terms across instances. G-Eval metrics likewise attain low CRPS, but their correlation is consistently low, suggesting a different regime: predictions match typical score levels (often because true scores are tightly concentrated), yet are less informative for distinguishing instances.

In contrast, LLMEval, BiGGen Judge, GEMBA, and FactScore produce broader target distributions, leading to slightly higher CRPS while still maintaining relatively high correlation. This pattern suggests that these metrics contain more instance-to-instance variation in quality, enabling meaningful relative comparisons even when uncertainty-aware prediction is harder. Overall, different tasks and metrics exhibit different mixtures of (i) distributional predictability of absolute scores and (ii) relative, instance-discriminative predictability.

\input{table/direct_assessment}

\input{table/ac_bc_sample_table}

\input{images/training_data_size}

\section{Conclusion}
\label{sec:conclude}

In this work, we propose the task of \emph{Performance Interval Estimation} (PIE) for predicting quality of free-form LLM generations. While confidence estimation is well-known for predicting the probability for whether model outputs are correct or not, predicting binary correctness in this way is most sensible when generation tasks have exact answers. In contrast, free-form generations are often more nuanced, with output quality being both fine-grained and multi-faceted. 

We therefore motivate prediction of arbitrary continuous, reference-based metric scores (point estimates) with calibrated prediction intervals. Framing this prediction task as regression allows us to be agnostic as to both: 1) whether the prediction target is ordinal or continuous; and 2) whether we are predicting an observed value (e.g., human rating or program execution time) or a derived evaluation metric. We emphasize the importance of prediction intervals for risk-informed decision-making; a confidence interval lower-bounds expected quality to better assess the level of risk. 

To promote this prediction task and foster community progress, we release our dataset, PIE-data, spanning 11 datasets across summarization, machine translation, code generation, function-calling, and question-answering, with multiple task evaluation metrics to predict. We also share our implemented methods and their outputs.  

Our experiments compare two alternative prediction methods: confidence-based regression (CE-Reg), which extracts off-the-shelf verbalized and consistency-based confidence signals and trains classic regression models to output both point estimates and prediction intervals; and reference-free LLM-as-a-judge (RF-LLMaaJ), which prompts an LLM with in-context learning to yield metric estimates and 95\% intervals for each input-output-instance.  CE-Reg is substantially more accurate and better calibrated than RF-LLMaaJ, especially with consistency features, showing that lightweight supervised regression over simple confidence signals can outperform few-shot LLM judging for both prediction and uncertainty quantification.

\section*{Acknowledgments}
This research was supported in part by Cisco, Amazon, and by Good Systems\footnote{\url{https://goodsystems.utexas.edu/}} (a UT Austin Grand Challenge dedicated to developing responsible AI technologies), as well as CosmicAI\footnote{The NSF-Simons AI Institute for Cosmic Origins (CosmicAI): \url{https://www.cosmicai.org/}} support from the NSF (Cooperative Agreement 2421782) and the Simons Foundation (grant MPS-AI-00010515). We thank the Texas Advanced Computing Center (TACC) for use of its superb infrastructure. The statements made herein are solely the opinions of the authors and do not reflect the views of the sponsoring agencies.

\bibliography{tacl2021}
\bibliographystyle{acl_natbib}

\newpage
\appendix
\section{Appendix}

\subsection{Reproducibility Details \label{sec:appendix_prompt}} 
\textbf{Training Regression Model}
Training regression models requires a set of annotated data for supervision. For a given task, a practitioner should produce a \textit{training set} by determining a set of $N$ questions and gold standard responses to those questions. Then an LLM of the practitioner's choosing should generate responses to those questions, which are compared against the gold standard responses using a task-specific evaluation metric (such as BERTScore \citep{bert-score}) -- resulting in a task-specific score $s$ for each question. Additionally, the practitioner should use whatever confidence estimation methods (at their discretion) to produce confidence estimates $C$ for each question. The next step is to train a regression model of choice on this training set of $\mathcal{D}$ and $C$ over $N$ questions. This model can subsequently be used to yield a predictive distribution around the task-specific metric score for any new LLM responses to out-of-sample questions.

\subsection{Sample Prompts}
\label{app:sampleprompts}



\textbf{Prompts for Confidence (Verbalized 1S)} \citep{tian-etal-2023-just}
\begin{Verbatim}[breaklines=true, frame=single,fontsize=\small] 
Questions: {question}

Please provide your answer to the following question along with your confidence level (0% to 100%). 
Respond **using only** your answer and confidence level, without any additional explanation.\n
Format your response **strictly as JSON** in this exact format:\n
'{"Answer": "Your answer as a string", "Confidence": Your confidence level as a number}'
)
\end{Verbatim}
    
\textbf{Prompts for Confidence (Verbalized 2S)} \citep{tian-etal-2023-just}
\begin{Verbatim}[breaklines=true, frame=single,fontsize=\small] 
Question: {question}

Provide the probability that each of your guesses is correct. Give ONLY
the probabilities, no other words or explanation. The probabilities for each guess are independent.\n\n 
Please answer question using JSON format: {'P1': <the probability between 0.0 and 1.0 that G1 is correct, without any extra commentary whatsoever; just the probability!>, ..., 'Pk': P${k}: <the probability between 0.0 and 1.0 that G${k} is correct, without any extra commentary whatsoever; just the probability!>}\n
\end{Verbatim}

\textbf{Prompts for RFLLMaaJ Baseline} 
\begin{Verbatim}[breaklines=true, frame=single,fontsize=\small] 
## Task ##\n
Given several examples of system input-output pairs along with their scores, predict what score the following test input-output pair would receive on the same metric.\n\n
## Shared Task Context ##\n
{task_context}\n\n
## Scoring Metric ##\n
"The scores are based on the metric: {metric}. This metric may originate from an automatic measure 
"(e.g., ROUGE, BLEU) or a model-based evaluation.\n\n
## Scored Examples ##\n
{examples_str}\n\n
## Test Pair to Score ##\n
System Input:\n{test_input}\n\n
System Output:\n{test_response}\n\n
## Required Output Format ##\n
Provide your prediction in the following JSON format:\n
{\n"
'    "most_likely_estimate": x,\n'
'    "lower_bound_at_95": xL,\n'
'    "upper_bound_at_95": xU\n'
"}          
\end{Verbatim}

\textbf{Prompts for LLM-based Evaluation Metrics (G-Eval)} \citep{liu-etal-2023-g}

\begin{Verbatim}[breaklines=true, frame=single,fontsize=\small] 
You will be given a news article. You will then be given one summary written for this article.

Your task is to rate the summary on one metric.

Please make sure you read and understand these instructions carefully. 
Please keep this document open while reviewing, and refer to it as needed.

Evaluation Criteria:

{evaluation_instructions}

Evaluation Steps:

1. Read the news article carefully and identify the main facts and key points.
2. Read the summary and compare it to the article. Check if the summary meets the {metric} criterion.
3. Assign a score for {metric} based on the Evaluation Criteria.

Source Text:

{source}

Summary:

{prediction}

Evaluation Form (scores ONLY), provide your response in the following format:
- {metric}:

evaluation_criteria = {
    "Coherence": """Coherence (1-5) - the collective quality of all sentences. A well-structured summary should build from sentence to a coherent body of information about a topic.""",
    "Consistency": """Consistency (1-5) - the factual alignment between the summary and the 
    source document. A factually consistent summary contains only statements entailed by the source document and avoids hallucinated facts.""",
    "Fluency": """Fluency (1-3): the quality of the summary in terms of grammar, spelling, punctuation, word choice, and sentence structure.

    - 1: Poor. The summary has many errors that make it hard to understand or sound unnatural.
    - 2: Fair. The summary has some errors that affect the clarity or smoothness of the text, but the main points are still comprehensible.
    - 3: Good. The summary has few or no errors and is easy to read and follow.""", 
    "Relevance": """Relevance (1-5) - selection of important content from the source. The summary should include only important information from the source document. Annotators were instructed to penalize summaries which contained redundancies and excess information."""
}
\end{Verbatim}

\textbf{Prompts for LLM-based Evaluation Metrics (GEMBA\_ref)} \citep{kocmi-federmann-2023-large}
\begin{Verbatim}[breaklines=true, frame=single,fontsize=\small] 
Evaluate the quality of the following translation from {source_lang} to {target_lang} on a scale from 0 to 100.\n

The scale starts from 'No meaning preserved', goes through 'Some meaning preserved', 'Most meaning preserved with few grammar mistakes', and up to 'Perfect meaning and grammar'.\n\n

Source Text (in {source_lang}):\n{source}\n\n

Model's Translation (in {target_lang}):\n{prediction}\n\n

Human Reference Translations (in {target_lang}): \n 

{reference_texts}\n\n

Provide your response in the following format: Score:
                
\end{Verbatim}

\textbf{Prompts for LLM-based Evaluation Metrics (LLM-Eval for QA)} 
\citep{10.1145/3726302.3729895, lin-chen-2023-llm}
\begin{Verbatim}[breaklines=true, frame=single,fontsize=\small] 
Task: Score the following LLM output of a question-answering task with respect to the following aspects using a 1 to 5 star rating system.
    
    Dataset: The dataset is a Question-Answering dataset, specifically designed for evaluating factual precision and detailed comparative reasoning in AI-generated answers.
    
    Output: Begin your evaluation by providing a short explanation. Be as objective as possible. After your explanation, provide your scores in JSON format like: [[SCORE]] {{"accuracy": 2, "informativeness": 3}}
    
    Criteria:
    - Accuracy:
        1 star: Incorrect information
        2 stars: Partially correct information
        3 stars: Half correct information
        4 stars: Mostly correct information
        5 stars: Perfectly correct information
    
    - Informativeness:
        1 star: No or irrelevant information
        2 stars: Very little information
        3 stars: Some information
        4 stars: Enough information
        5 stars: Highly informative
    
    Question:
    {source}
    
    Provided Answer:
    {ans}
    
    Reference Answer(s):
    {ref_text}
    
    Evaluation:
    [[SCORE]] {{\"accuracy\": <1–5>, \"informativeness\": <1–5>}}
\end{Verbatim}

\textbf{Prompts for LLM-based Evaluation Metrics (LLM-as-a-Judge for BiGGen)} \citep{kim-etal-2025-biggen}

System Prompt:
\begin{Verbatim}[breaklines=true, frame=single,fontsize=\small] 
You are a fair judge assistant tasked with providing clear, objective feedback based on specific criteria, ensuring each assessment reflects the absolute standards set for performance.
\end{Verbatim}

User Prompt:
\begin{Verbatim}[breaklines=true, frame=single,fontsize=\small] 
###Task Description:
An instruction (might include an Input inside it), a response to evaluate, and a score rubric representing an evaluation criteria are given.
1. Write a detailed feedback that assesses the quality of the response strictly based on the given score rubric, not evaluating in general.
2. After writing a feedback, write a score that is an integer between 1 and 5. You should refer to the score rubric.
3. The output format should look as follows: "(write a feedback for criteria) [RESULT] (an integer number between 1 and 5)"
4. Please do not generate any other opening, closing, and explanations.

###The instruction to evaluate:
{instruction}

###Response to evaluate:
{response}

###Score Rubrics:
{rubric}

###Feedback: 
\end{Verbatim}

\textbf{Prompts for LLM-based Evaluation Metrics (CodeJudge for code generation and function-calling)} 
\cite{tong-zhang-2024-codejudge}

Functional Correctness:
\begin{Verbatim}[breaklines=true, frame=single,fontsize=\small] 
You will be given the code snippet for a problem.
Your task is to rate the code snippet only on one metric.\n\n
Evaluation Criteria:\n
Functional Correctness (0-4): Execution-based quality of the code snippet combined with the problem.\n
- 0: Fails all tests, totally incorrect.\n
- 4: Passes all tests, totally correct.\n
{instr}\nProblem:\n{source}\n\n
Reference Code:\n{reference}\n\nCode Snippet:\n{answer}\n\n
Evaluation Form:\nFunctional Correctness (scores ONLY):                
\end{Verbatim}

Inconsistency Prompt:
\begin{Verbatim}[breaklines=true, frame=single,fontsize=\small] 
You will be provided with a problem statement, a code snippet that supposedly addresses the problem,
and a catalog of code inconsistencies.
Evaluation Steps:
1. Read the problem statement carefully to identify the functionalities required for the
implementation.
2. Read the code snippet and compare it to the problem statement. Check if the code snippet covers
the required functionalities.
3. Output your answer in a JSON format list.
a) If the code snippet is correct, output: ["inconsistency": "None", "severity": "Negligible"].
b) If the code snippet is incorrect, output the identified inconsistencies and their severity
according to the catalog of code inconsistencies. For example: ["inconsistency": "<inconsistency1>",
"severity": "<severity1>", "inconsistency": "<inconsistency2>", "severity": "<severity2>", ...]
Problem: {PROBLEM}
Code Snippet: {CODE}
Taxonomy of Common Inconsistencies:
1. Missing dependency declarations: Negligible
2. No error messages for unexpected input cases: Negligible
3. Inefficiency, unnecessary statements: Negligible
4. Edge case not handled: Small
5. Logic error: Major
6. Function or variable not defined: Fatal
7. Code not completed: Fatal
Evaluation Form:
JSON output (a JSON list only):
\end{Verbatim}

\subsection{Additional Dataset Information}
\input{table/data_analysis}

\input{table/favorite_target_metric}
\newpage
\subsection{Complementary Tables and Figures}

\input{images/confidence_interval_radar_plot_}

\input{table/ac_bc_table}

\input{table/ac_bc_table_corr}

\input{table/ac_bc_table_ace}









\end{document}

%% file: table/ac_bc_table_crps.tex
\begin{table*}[!t]
  \centering
  \makebox[\textwidth][l]{%
    \hspace*{-0.5cm}%
  \scriptsize
  \setlength{\tabcolsep}{4pt}
  \renewcommand{\arraystretch}{1.2}
    \begin{tabular}{|
        l|l|
        *{5}{c}|
        *{2}{>{\centering\arraybackslash}p{1.2cm}}|
        *{3}{>{\centering\arraybackslash}p{1.0cm}}|
      }
      \midrule
      \multicolumn{2}{|c|}{} 
        & \multicolumn{5}{c|}{\textbf{CE$-$Reg: Consistency-based}}
        & \multicolumn{2}{c|}{\textbf{CE$-$Reg: Verbalized}}
        & \multicolumn{3}{c|}{\textbf{RF-LLMaaJ}} \\
      \cline{3-12}
      \textbf{Dataset} 
        & \textbf{Metric} 
        & \textbf{DegMat.} & \textbf{Eccen.} & \textbf{EigVal.} 
        & \textbf{LexSim.} & \textbf{NumSet.} 
        & \textbf{1S}   & \textbf{2S} 
        & \textbf{4-shot} & \textbf{8-shot} & \textbf{16-shot} \\
      \midrule
      \rowcolor{gray!10}
        & Judge(Func.)  & 0.22 & 0.27 & 0.28 & 0.25 & 0.33 
                        & 0.36 & 0.34 
                        & 0.28 & 0.27 & 0.26 \\
      \rowcolor{gray!10}
        & Judge(Incon.) & 0.05 & 0.09 & 0.09 & 0.08 & 0.11 
                        & 0.12 & 0.11 
                        & 0.14 & 0.14 & 0.13 \\
      \rowcolor{gray!10}
      \multirow{-3}{*}{APIGen}
        & CodeBLEU      & 0.02 & 0.04 & 0.04 & 0.03 & 0.05 
                        & 0.05 & 0.05 
                        & 0.06 & 0.06 & 0.06 \\
      \midrule
      \multirow{3}{*}{ASQA}
        & LLMEval(A.)   & 0.14 & 0.18 & 0.13 & 0.13 & 0.20 
                        & 0.18 & 0.19 
                        & 0.16 & 0.16 & 0.16 \\
        & LLMEval(I.)   & 0.09 & 0.13 & 0.09 & 0.09 & 0.14 
                        & 0.14 & 0.14 
                        & 0.13 & 0.11 & 0.13 \\
      \multirow{-3}{*}{ASQA}
        & BERTScore     & 0.02 & 0.02 & 0.02 & 0.01 & 0.02 
                        & 0.02 & 0.02 
                        & 0.03 & 0.02 & 0.02 \\
      \midrule
      \rowcolor{gray!10}
      \multirow{-1}{*}{BiGGen}
        & Judge         & 0.17 & 0.26 & 0.18 & 0.17 & 0.26 
                        & 0.26 & 0.24 
                        & 0.25 & 0.26 & 0.25 \\
      \midrule
      \multirow{3}{*}{BigCodeBench}
        & Judge(Func.)  & 0.21 & 0.24 & 0.22 & 0.22 & 0.29 
                        & 0.29 & 0.28 
                        & 0.35 & 0.37 & 0.38 \\
        & Judge(Incon.) & 0.03 & 0.04 & 0.04 & 0.04 & 0.07 
                        & 0.07 & 0.06 
                        & 0.12 & 0.10 & 0.09 \\
      \multirow{-3}{*}{BigCodeBench}
        & CodeBLEU      & 0.05 & 0.06 & 0.05 & 0.05 & 0.07 
                        & 0.07 & 0.07 
                        & 0.09 & 0.08 & 0.08 \\
      \midrule
      \rowcolor{gray!10}
        & LLMEval(A.)   & 0.15 & 0.25 & 0.15 & 0.15 & 0.24 
                        & 0.25 & 0.23 
                        & 0.23 & 0.24 & 0.23 \\
      \rowcolor{gray!10}
        & LLMEval(I.)   & 0.10 & 0.14 & 0.10 & 0.10 & 0.13 
                        & 0.14 & 0.13 
                        & 0.14 & 0.13 & 0.13 \\
      \rowcolor{gray!10}
      \multirow{-3}{*}{Eli5}
        & BERTScore     & 0.01 & 0.01 & 0.01 & 0.01 & 0.01 
                        & 0.01 & 0.01 
                        & 0.02 & 0.02 & 0.02 \\
      \midrule
      \multirow{-1}{*}{Fact-Bio}
        & FActScore     & 0.10 & 0.26 & 0.10 & 0.11 & 0.26 
                        & 0.19 & 0.26 
                        & 0.23 & 0.21 & 0.22 \\
      \midrule
      \rowcolor{gray!10}
        & LLMEval(A.)   & 0.16 & 0.25 & 0.16 & 0.16 & 0.25 
                        & 0.25 & 0.25 
                        & 0.23 & 0.23 & 0.22 \\
      \rowcolor{gray!10}
        & LLMEval(I.)   & 0.11 & 0.13 & 0.11 & 0.11 & 0.14 
                        & 0.14 & 0.13 
                        & 0.14 & 0.13 & 0.13 \\
      \rowcolor{gray!10}
      \multirow{-3}{*}{MED LF QA}
        & BERTScore     & 0.02 & 0.02 & 0.02 & 0.01 & 0.02 
                        & 0.02 & 0.02 
                        & 0.03 & 0.03 & 0.03 \\
      \midrule
      \multirow{2}{*}{Opus-100}
        & GEMBA         & 0.10 & 0.14 & 0.11 & 0.11 & 0.17 
                        & 0.16 & 0.16 
                        & 0.16 & 0.14 & 0.14 \\
      \multirow{-2}{*}{Opus-100}
        & BERTScore     & 0.03 & 0.05 & 0.04 & 0.03 & 0.06 
                        & 0.06 & 0.06 
                        & 0.07 & 0.07 & 0.07 \\
      \midrule
      \rowcolor{gray!10}
        & G-Eval(Coh.)  & 0.04 & 0.05 & 0.04 & 0.04 & 0.05 
                        & 0.05 & 0.05 
                        & 0.08 & 0.07 & 0.08 \\
      \rowcolor{gray!10}
        & G-Eval(Con.)  & 0.06 & 0.08 & 0.05 & 0.06 & 0.08 
                        & 0.08 & 0.07 
                        & 0.10 & 0.09 & 0.10 \\
      \rowcolor{gray!10}
        & G-Eval(Flu.)  & 0.01 & 0.02 & 0.01 & 0.01 & 0.02 
                        & 0.02 & 0.02 
                        & 0.03 & 0.03 & 0.03 \\
      \rowcolor{gray!10}
        & G-Eval(Rel.)  & 0.11 & 0.14 & 0.11 & 0.11 & 0.14 
                        & 0.14 & 0.14 
                        & 0.14 & 0.14 & 0.14 \\
      \rowcolor{gray!10}
      \multirow{-5}{*}{Summeval}
        & BERTScore     & 0.01 & 0.01 & 0.01 & 0.01 & 0.01 
                        & 0.01 & 0.01 
                        & 0.02 & 0.02 & 0.02 \\
      \midrule
        & LLMEval(A.)   & 0.16 & 0.29 & 0.16 & 0.16 & 0.31 
                      & 0.29 & 0.28 
                      & 0.30 & 0.30 & 0.30 \\
      & LLMEval(I.)   & 0.16 & 0.23 & 0.17 & 0.16 & 0.25 
                      & 0.24 & 0.23 
                      & 0.25 & 0.25 & 0.24 \\
    \multirow{-3}{*}{TruthfulQA}
      & BERTScore     & 0.02 & 0.03 & 0.02 & 0.02 & 0.03 
                      & 0.03 & 0.03 
                      & 0.06 & 0.05 & 0.05 \\
      \midrule
      \multicolumn{2}{|l|}{AVERAGE}
        & \textbf{0.09} & 0.13 & \textbf{0.09} & \textbf{0.09} & 0.14 
        & 0.13 & 0.13 
        & 0.14 & 0.14 & 0.14 \\
      \bottomrule
    \end{tabular}
  }
  \caption{\textbf{CRPS Benchmark.} We compare two baselines: 1) confidence-based regression (CE-Reg) uses 7 CE features (five consistency-based, two verbalized); and 2) reference-free LLM-as-a-Judge (RF-LLMaaJ), evaluated in 4-, 8-, and 16-shot settings. CE-Reg generally scores better (lower) CRPS than RF-LLMaaJ.}
  \label{tab:benchmark_crps}
\end{table*}

%% file: images/model_compare_fig.tex
\begin{figure*}[t]
    \centering
    \includegraphics[width=\textwidth]{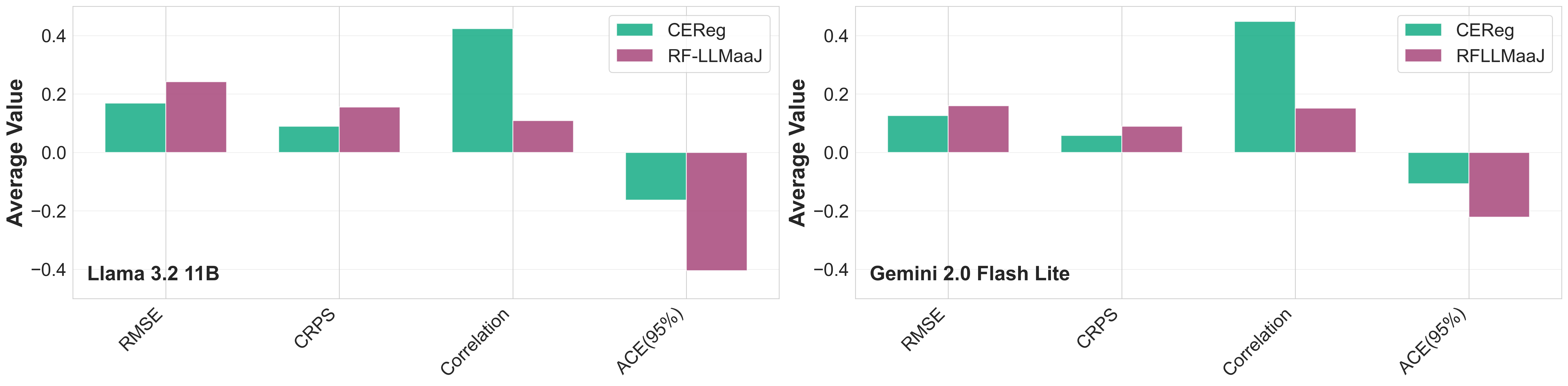}
    \caption{CE-Reg outperforms RF-LLMaaJ on RMSE, CRPS, Correlation, and ACE (95\%) on Llama 3.2 11B (left) and Gemini 2.0 flash-lite (right), averaged across all datasets and task metrics.
    Lower is better for all except correlation. This result shows minimal difference between LLMs.
    }
    \label{fig:model_comparison_ce_vs_rf}
\end{figure*}

%% file: table/direct_assessment.tex
\begin{table}[!t]
  \centering
  \scriptsize
  \setlength{\tabcolsep}{4pt}
  \renewcommand{\arraystretch}{1.2}
  \begin{tabular}{|c|c|ccc|}
    \hline
    \multirow{2}{*}{
    }
      & \multicolumn{1}{c|}{\textbf{CE-Reg: Verbalized}} 
      & \multicolumn{3}{c|}{\textbf{RF-LLMaaJ}} \\
    \cline{2-5}
    & \textbf{2S} 
      & \textbf{4-shot} & \textbf{8-shot} & \textbf{16-shot} \\
    \hline
    CRPS       & 0.21 & 0.22 & 0.22	&0.24 \\
    \hline
  \end{tabular}
  \caption{\textbf{Direct Assessment:} For BiGGen we evaluate two baselines using pre-generated outputs—Verbalized 2S and RF-LLMaaJ—for directly predicting human ratings. 
  Methods requiring additional outputs are omitted. This result shows CE-Reg calibrating better than LLMaaJ on predicting human judgments.
  } 
  \label{tab:direct_assessment}
\end{table}

%% file: table/ac_bc_sample_table.tex

\begin{table}[!t]
  \centering
  \scriptsize
  \setlength{\tabcolsep}{4pt}
  \renewcommand{\arraystretch}{1.2}
  \begin{tabular}{|c|ccc|ccc|}
    \hline
    \multirow{2}{*}{
    }
      & \multicolumn{3}{c|}{\textbf{CE-Reg: DegMat}} 
      & \multicolumn{3}{c|}{\textbf{RF-LLMaaJ}} \\
    \cline{2-7}
    & \textbf{4 inst.} & \textbf{8 inst.} & \textbf{16 inst.} 
      & \textbf{4-shot} & \textbf{8-shot} & \textbf{16-shot}\\
    \hline
    
    CRPS       & 0.09  & 0.09 & 0.09 & 0.14	& 0.14 & 0.14 \\
    \hline
  \end{tabular}
  \caption{Comparing
 CE-Reg using DegMat feature vs.\ RF-LLMaaJ, varying training example count, averaged across all datasets, showing undetectable benefit from additional training examples on CRPS.}
  \label{tab:sample_table}
\end{table}

%% file: images/training_data_size.tex
\begin{figure}[!ht]
    \centering
    \includegraphics[width=0.49\textwidth]{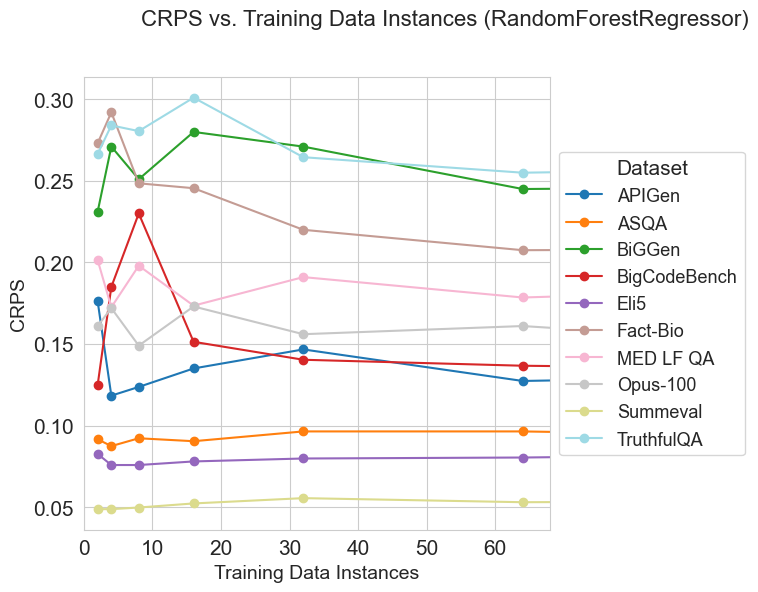}
    \caption{Impact of training data size on CE-Reg performance, measured by test-set CRPS across datasets using the selected task-specific metric (see Table~\ref{tab:selected_targets}). Training data sizes are capped at 64 instances. CRPS stabilizes after approximately 16 instances, suggesting a 
    plateau for certain datasets.}
    \label{fig:train_size_plot}
\end{figure}

%% file: table/data_analysis.tex
\begin{table*}[!t]
  {\rowcolors{2}{gray!25}{white}%
  \centering
  \scriptsize
  \renewcommand{\arraystretch}{1.2}
  \begin{tabularx}{\textwidth}{|>{\raggedright\arraybackslash}p{2.5cm}|l|r|X|}
    \toprule
    \textbf{Dataset} & \textbf{Type} & \textbf{Data Split Size} & \textbf{Task-Specific Metrics} \\
    \midrule
    APIGen \citep{10.5555/3737916.3739641} 
      & Function Calling 
      & 8955 / 2985 / 2985 
      & CodeBLEU, CodeJudge (functional correctness, inconsistency)[gemini-2.5-flash]\\[0.5ex]
    ASQA \citep{stelmakh-etal-2022-asqa} 
      & Question Answering 
      & 1689 / 563 / 564  
      & LLM‑EVAL (accuracy, informativeness) [gemini-2.5-flash], BERTScore \\[0.5ex]
    BigCodeBench \citep{zhuo2025bigcodebench} 
      & Code Generation 
      & 1994	/ 665	/ 665 
      & CodeBLEU, CodeJudge (functional correctness, inconsistency)[gemini-2.5-flash]\\[0.5ex]
    BiGGen \citep{kim-etal-2025-biggen} 
      & Open-Ended Generation 
      & 847 / 283 / 283 
      & BiGGen-Judge [gemini-2.5-flash] \\[0.5ex]
    BiGGen-Bench-Results \citep{kim-etal-2025-biggen} 
      & Open-Ended Generation 
      & 1662 / 555 / 555
      & Human Rating \\[0.5ex]
    Eli5 \citep{fan-etal-2019-eli5} 
      & Question Answering 
      & 2657 / 886 / 886 
      & BERTScore, LLM‑EVAL (accuracy, informativeness)  [gemini-2.5-flash] \\[0.5ex]
    Fact-Bio \citep{min-etal-2023-factscore} 
      & Fact-Checking 
      & 1216 / 406 / 406 
      & FActScore [gpt-3.5-turbo] \\[0.5ex]
    Med LF QA \citep{hosseinibenchmark} 
      & Medical QA 
      & 8855 / 2952 / 2952 
      & BERTScore, LLM‑EVAL (accuracy, informativeness) [gemini-2.5-flash]  \\[0.5ex]
    Opus-100 (EN→ES) \citep{zhang-etal-2020-improving,tiedemann-2012-parallel} 
      & Machine Translation 
      & 3579 / 1194 / 1194 
      & BERTScore, GEMBA\_SQM\_ref [gemini-2.5-flash] \\[0.5ex]
    Summeval \citep{fabbri2021summeval} 
      & Text Summarization 
      & 2809 / 937 / 937 
      & BERTScore, G-Eval (Coherence, Consistency, Fluency, Relevance) [gemini-2.5-flash] \\[0.5ex]
    TruthfulQA \citep{lin-etal-2022-truthfulqa} 
      & Question Answering 
      & 1455 / 486 / 486 
      & BERTScore, LLM‑EVAL (accuracy, informativeness) [gemini-2.5-flash] \\
    \bottomrule
  \end{tabularx}
  }
  \caption{Our study and benchmark cover twelve datasets spanning a range of diverse tasks: code generation, fact-checking, function-calling, machine translation (MT), question-answering (QA), summarization, and others. The original dataset sizes is capped at 5,000 instances (through random sampling), with each instances generating three outputs, we collect maximum 15,000 output-score instance per datasets. Train-dev-test is split by 3:1:1. For each dataset, we also specify one or more task-specific, multi-dimensional and partial-credit evaluation metric used to evaluate generation performance. BiGGen-Bench-Results is the direct assessment evaluation setting.}
  \label{tab:dataset_summary}
\end{table*}

%% file: table/favorite_target_metric.tex
{{\rowcolors{2}{gray!25}{white}
\begin{table*}[!ht]

\centering
\small
\renewcommand{\arraystretch}{1.2}
\begin{tabular}{|l|l|}
\hline
\textbf{Dataset} & \textbf{Selected Metrics} \\
\hline
APIGen \citep{10.5555/3737916.3739641} & Judge (Funct.) \\
ASQA \citep{stelmakh-etal-2022-asqa} & LLMEval(A.) \\
BigCodeBench \citep{zhuo2025bigcodebench} & Judge (Funct.) \\
BiGGen \citep{kim-etal-2025-biggen} & Judge \\
Eli5 \citep{fan-etal-2019-eli5} & LLMEval(A.)  \\
Fact-Bio \citep{min-etal-2023-factscore} & FActScore \\
MED LF QA \citep{hosseinibenchmark} & LLMEval(A.)  \\
Opus-100 \citep{zhang-etal-2020-improving, tiedemann-2012-parallel} & GEMBA  \\
Summeval \citep{fabbri2021summeval} & G-Eval(Rel.) \\
TruthfulQA \citep{lin-etal-2022-truthfulqa} & LLMEval(A.)  \\
\hline
\end{tabular}
\caption{While Table~\ref{tab:dataset_summary} includes our full set of task-specific evaluation metrics, for brevity we sometimes select only a single metric per dataset to report, as listed in this table here. These selected metrics are the ones reported on in Figure \ref{fig:train_size_plot}. 
}
\label{tab:selected_targets}
\end{table*}
}
}

%% file: images/confidence_interval_radar_plot_.tex
\begin{figure}[!htb]
  \centering

  \begin{subfigure}[b]{0.75\linewidth}
    \centering
    \includegraphics[width=\textwidth]{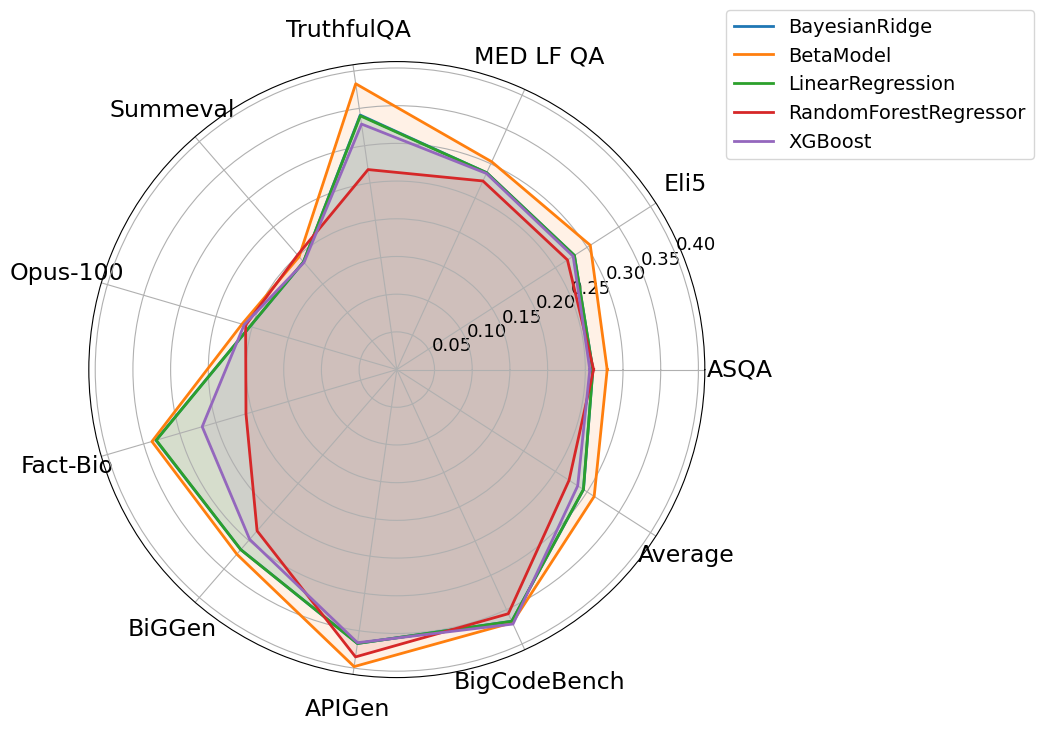}
    \caption{RMSE for regression models}
    \label{fig:rmse}
  \end{subfigure}

  \vspace{1em}  

  \begin{subfigure}[b]{0.75\linewidth}
    \centering
    \includegraphics[width=\textwidth]{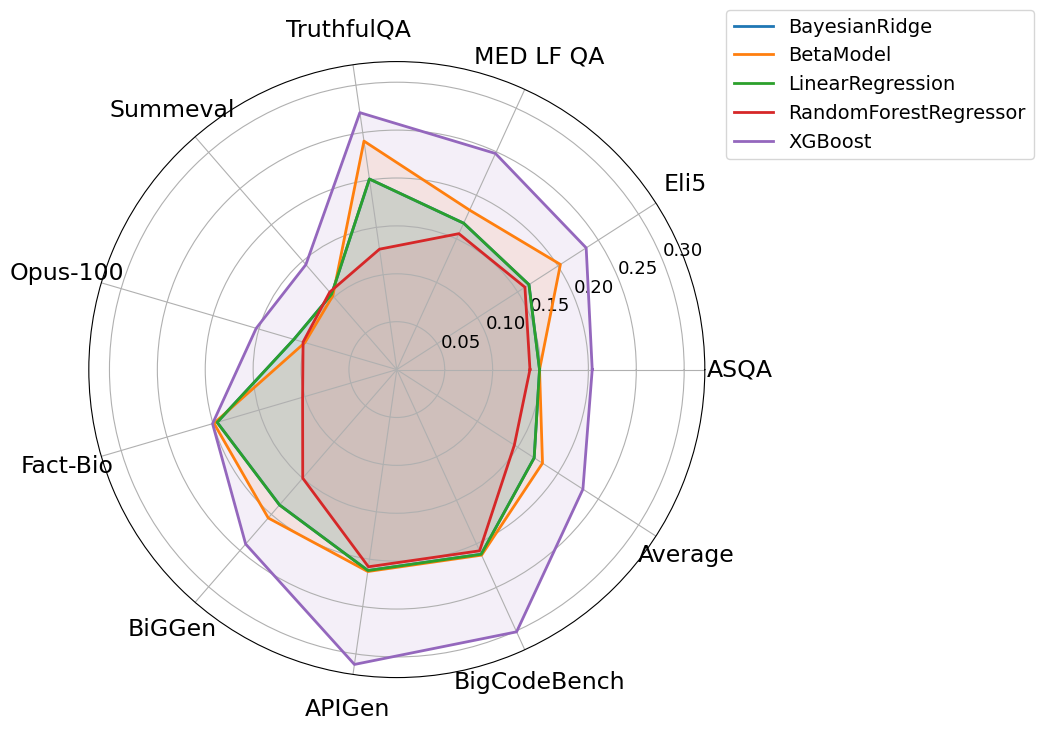}
    \caption{CRPS for regression models}
    \label{fig:crps}
  \end{subfigure}

  \vspace{1em}
  
  \begin{subfigure}[b]{0.75\linewidth}
    \centering
    \includegraphics[width=\textwidth]{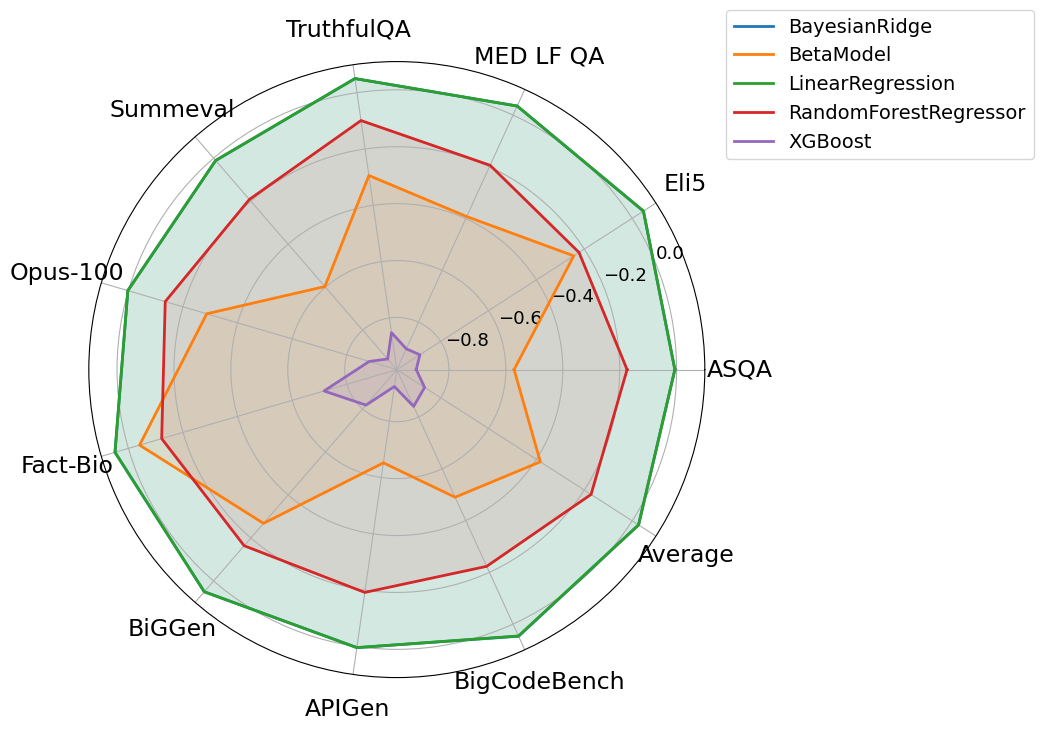}
    \caption{ACE(95\%) for regression models}
    \label{fig:ace}
  \end{subfigure}

  \vspace{1em}

  \begin{subfigure}[b]{0.75\linewidth}
    \centering
    \includegraphics[width=\textwidth]{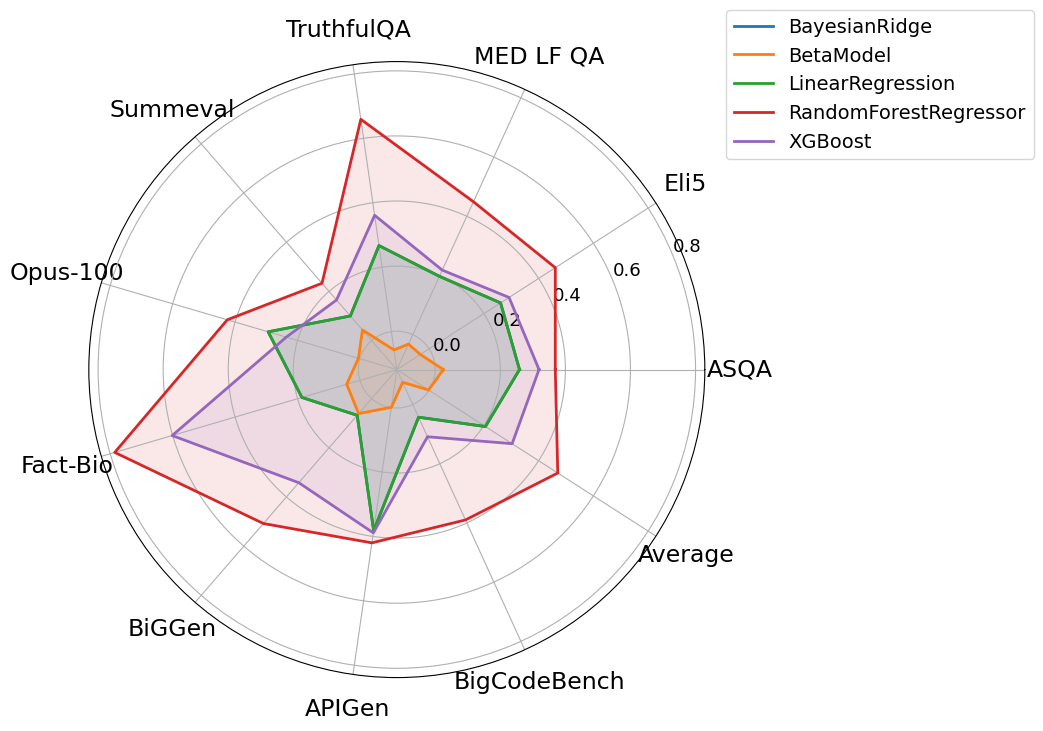}
    \caption{Pearson correlation for regression models}
    \label{fig:corr}
  \end{subfigure}



  \caption{Comparison across regression models on development sets. For CRPS, RMSE, ACE(95\%) lower is better; for Pearson correlation higher is better.}
  \label{fig:radar_plots}
\end{figure}

%% file: table/ac_bc_table.tex
\begin{table*}[!t]
  \centering
  \makebox[\textwidth][l]{%
    \hspace*{-0.5cm}%
  \scriptsize
  \setlength{\tabcolsep}{4pt}
  \renewcommand{\arraystretch}{1.2}
    \begin{tabular}{|
        l|l|
        *{5}{c}|
        *{2}{>{\centering\arraybackslash}p{1.2cm}}|
        *{3}{>{\centering\arraybackslash}p{1.0cm}}|
      }
      \midrule
      \multicolumn{2}{|c|}{} 
        & \multicolumn{5}{c|}{\textbf{CE$-$Reg: Consistency-based}}
        & \multicolumn{2}{c|}{\textbf{CE$-$Reg: Verbalized}}
        & \multicolumn{3}{c|}{\textbf{RF-LLMaaJ}} \\
      \cline{3-12}
      \textbf{Dataset} 
        & \textbf{Metric} 
        & \textbf{DegMat.} & \textbf{Eccen.} & \textbf{EigVal.} 
        & \textbf{LexSim.} & \textbf{NumSet.} 
        & \textbf{1S}   & \textbf{2S} 
        & \textbf{4-shot} & \textbf{8-shot} & \textbf{16-shot} \\
      \midrule
      \rowcolor{gray!10}
        & Judge(Func.)  & 0.39 & 0.38 & 0.40 & 0.39 & 0.37 
                        & 0.39 & 0.39 
                        & 0.43 & 0.43 & 0.43 \\
      \rowcolor{gray!10}
        & Judge(Incon.) & 0.11 & 0.12 & 0.12 & 0.12 & 0.12 
                        & 0.12 & 0.12 
                        & 0.23 & 0.21 & 0.20 \\
      \rowcolor{gray!10}
      \multirow{-3}{*}{APIGen}
        & CodeBLEU      & 0.05 & 0.06 & 0.06 & 0.06 & 0.06 
                        & 0.06 & 0.06 
                        & 0.12 & 0.11 & 0.10 \\
      \midrule
      \multirow{3}{*}{ASQA}
        & LLMEval(A.)   & 0.25 & 0.24 & 0.25 & 0.25 & 0.23 
                        & 0.23 & 0.24 
                        & 0.26 & 0.26 & 0.25 \\
        & LLMEval(I.)   & 0.17 & 0.17 & 0.17 & 0.17 & 0.17 
                        & 0.17 & 0.18 
                        & 0.20 & 0.18 & 0.20 \\
      \multirow{-3}{*}{ASQA}
        & BERTScore     & 0.02 & 0.02 & 0.02 & 0.02 & 0.03 
                        & 0.02 & 0.02 
                        & 0.08 & 0.05 & 0.05 \\
      \midrule
      \rowcolor{gray!10}
      \multirow{-1}{*}{BiGGen}
        & Judge         & 0.30 & 0.31 & 0.29 & 0.28 & 0.30 
                        & 0.30 & 0.31 
                        & 0.35 & 0.36 & 0.35 \\
      \midrule
      \multirow{3}{*}{BigCodeBench}
        & Judge(Func.)  & 0.36 & 0.34 & 0.36 & 0.37 & 0.35 
                        & 0.35 & 0.35 
                        & 0.48 & 0.49 & 0.50 \\
        & Judge(Incon.) & 0.09 & 0.09 & 0.09 & 0.09 & 0.09 
                        & 0.09 & 0.09 
                        & 0.22 & 0.19 & 0.17 \\
      \multirow{-3}{*}{BigCodeBench}
        & CodeBLEU      & 0.08 & 0.09 & 0.08 & 0.08 & 0.10 
                        & 0.10 & 0.10 
                        & 0.14 & 0.13 & 0.13 \\
      \midrule
      \rowcolor{gray!10}
        & LLMEval(A.)   & 0.26 & 0.28 & 0.26 & 0.26 & 0.27 
                        & 0.28 & 0.28 
                        & 0.33 & 0.33 & 0.32 \\
      \rowcolor{gray!10}
        & LLMEval(I.)   & 0.17 & 0.18 & 0.16 & 0.17 & 0.18 
                        & 0.18 & 0.18 
                        & 0.21 & 0.21 & 0.21 \\
      \rowcolor{gray!10}
      \multirow{-3}{*}{Eli5}
        & BERTScore     & 0.01 & 0.01 & 0.01 & 0.01 & 0.01 
                        & 0.01 & 0.01 
                        & 0.06 & 0.07 & 0.05 \\
      \midrule
      \multirow{-1}{*}{Fact-Bio}
        & FActScore     & 0.20 & 0.31 & 0.20 & 0.22 & 0.31 
                        & 0.25 & 0.32 
                        & 0.33 & 0.31 & 0.34 \\
      \midrule
      \rowcolor{gray!10}
        & LLMEval(A.)   & 0.28 & 0.29 & 0.28 & 0.28 & 0.28 
                        & 0.28 & 0.29 
                        & 0.34 & 0.34 & 0.32 \\
      \rowcolor{gray!10}
        & LLMEval(I.)   & 0.18 & 0.19 & 0.18 & 0.18 & 0.19 
                        & 0.18 & 0.18 
                        & 0.21 & 0.21 & 0.21 \\
      \rowcolor{gray!10}
      \multirow{-3}{*}{MED LF QA}
        & BERTScore     & 0.03 & 0.03 & 0.03 & 0.03 & 0.03 
                        & 0.03 & 0.03 
                        & 0.09 & 0.09 & 0.09 \\
      \midrule
      \multirow{2}{*}{Opus-100}
        & GEMBA         & 0.21 & 0.22 & 0.22 & 0.21 & 0.22 
                        & 0.22 & 0.23 
                        & 0.26 & 0.24 & 0.24 \\
      \multirow{-2}{*}{Opus-100}
        & BERTScore     & 0.06 & 0.07 & 0.07 & 0.06 & 0.08 
                        & 0.08 & 0.08 
                        & 0.12 & 0.12 & 0.12 \\
      \midrule
      \rowcolor{gray!10}
        & G-Eval(Coh.)  & 0.14 & 0.12 & 0.13 & 0.14 & 0.12 
                        & 0.12 & 0.12 
                        & 0.16 & 0.16 & 0.16 \\
      \rowcolor{gray!10}
        & G-Eval(Con.)  & 0.18 & 0.15 & 0.17 & 0.17 & 0.14 
                        & 0.15 & 0.15 
                        & 0.19 & 0.18 & 0.20 \\
      \rowcolor{gray!10}
        & G-Eval(Flu.)  & 0.05 & 0.05 & 0.05 & 0.06 & 0.05 
                        & 0.05 & 0.05 
                        & 0.07 & 0.07 & 0.07 \\
      \rowcolor{gray!10}
        & G-Eval(Rel.)  & 0.20 & 0.18 & 0.20 & 0.21 & 0.18 
                        & 0.18 & 0.18 
                        & 0.21 & 0.22 & 0.22 \\
      \rowcolor{gray!10}
      \multirow{-5}{*}{Summeval}
        & BERTScore     & 0.02 & 0.02 & 0.02 & 0.02 & 0.02 
                        & 0.02 & 0.02 
                        & 0.06 & 0.06 & 0.06 \\
      \midrule
        & LLMEval(A.)   & 0.31 & 0.34 & 0.30 & 0.30 & 0.35 
                      & 0.34 & 0.34 
                      & 0.44 & 0.43 & 0.43 \\
      & LLMEval(I.)   & 0.27 & 0.28 & 0.27 & 0.26 & 0.29 
                      & 0.29 & 0.29 
                      & 0.35 & 0.36 & 0.34 \\
    \multirow{-3}{*}{TruthfulQA}
      & BERTScore     & 0.04 & 0.04 & 0.04 & 0.04 & 0.04 
                      & 0.04 & 0.04 
                      & 0.15 & 0.12 & 0.12 \\
      \midrule
      \multicolumn{2}{|l|}{AVERAGE}
        & \textbf{0.16} & 0.17 & \textbf{0.16} & \textbf{0.16} & 0.17 
        & 0.17 & 0.17 
        & 0.23 & 0.22 & 0.22 \\
      \bottomrule
    \end{tabular}
  }
  \caption{\textbf{RMSE Benchmark.} We compare two baselines: 1) confidence-based regression (CE-Reg) uses 7 CE features (five consistency-based, two verbalized); and 2) reference-free LLM-as-a-Judge (RF-LLMaaJ), evaluated in 4-, 8-, and 16-shot settings. For RMSE, lower is better. Judge(Func.) = CodeJudge Functional Correctness. Judge(Incon.) = CodeJudge Inconsistency. LLMEval(A.) = LLM Eval Accuracy. LLMEval(I.) = LLMEval Informativeness.}
  \label{tab:benchmark_rmse}
\end{table*}

%% file: table/ac_bc_table_corr.tex
\begin{table*}[!t]
  \centering
  \makebox[\textwidth][l]{%
    \hspace*{-0.5cm}%
  \scriptsize
  \setlength{\tabcolsep}{4pt}
  \renewcommand{\arraystretch}{1.2}
    \begin{tabular}{|
        l|l|
        *{5}{c}|
        *{2}{>{\centering\arraybackslash}p{1.2cm}}|
        *{3}{>{\centering\arraybackslash}p{1.0cm}}|
      }
      \midrule
      \multicolumn{2}{|c|}{} 
        & \multicolumn{5}{c|}{\textbf{CE$-$Reg: Consistency-based}}
        & \multicolumn{2}{c|}{\textbf{CE$-$Reg: Verbalized}}
        & \multicolumn{3}{c|}{\textbf{RF-LLMaaJ}} \\
      \cline{3-12}
      \textbf{Dataset} 
        & \textbf{Metric} 
        & \textbf{DegMat.} & \textbf{Eccen.} & \textbf{EigVal.} 
        & \textbf{LexSim.} & \textbf{NumSet.} 
        & \textbf{1S}   & \textbf{2S} 
        & \textbf{4-shot} & \textbf{8-shot} & \textbf{16-shot} \\
      \midrule
      \rowcolor{gray!10}
        & Judge(Func.)  & 0.40 & 0.36 & 0.31 & 0.36 & 0.36 
                        & 0.17 & 0.20 
                        & 0.29 & 0.31 & 0.33 \\
      \rowcolor{gray!10}
        & Judge(Incon.) & 0.46 & 0.31 & 0.29 & 0.34 & 0.26 
                        & 0.13 & 0.17 
                        & -0.14 & -0.14 & -0.10 \\
      \rowcolor{gray!10}
      \multirow{-3}{*}{APIGen}
        & CodeBLEU      & 0.59 & 0.33 & 0.38 & 0.41 & 0.10 
                        & -0.01 & 0.15 
                        & 0.08 & 0.05 & 0.06 \\
      \midrule
      \multirow{3}{*}{ASQA}
        & LLMEval(A.)   & 0.33 & 0.24 & 0.35 & 0.36 & 0.30 
                        & 0.33 & 0.15 
                        & 0.24 & 0.22 & 0.25 \\
        & LLMEval(I.)   & 0.39 & 0.27 & 0.38 & 0.41 & 0.30 
                        & 0.28 & 0.15 
                        & 0.18 & 0.28 & 0.14 \\
      \multirow{-3}{*}{ASQA}
        & BERTScore     & 0.51 & 0.34 & 0.48 & 0.57 & 0.10 
                        & 0.33 & 0.23 
                        & 0.04 & 0.07 & 0.11 \\
      \midrule
      \rowcolor{gray!10}
      \multirow{-1}{*}{BiGGen}
        & Judge         & 0.44 & 0.13 & 0.44 & 0.50 & 0.27 
                        & 0.24 & 0.21 
                        & 0.19 & 0.08 & 0.16 \\
      \midrule
      \multirow{3}{*}{BigCodeBench}
        & Judge(Func.)  & 0.34 & 0.32 & 0.33 & 0.33 & 0.05 
                        & 0.10 & 0.04 
                        & 0.07 & 0.10 & 0.08 \\
        & Judge(Incon.) & 0.36 & 0.35 & 0.35 & 0.35 & 0.10 
                        & 0.06 & -0.01 
                        & -0.08 & -0.08 & -0.08 \\
      \multirow{-3}{*}{BigCodeBench}
        & CodeBLEU      & 0.58 & 0.50 & 0.56 & 0.60 & 0.12 
                        & 0.05 & 0.02 
                        & 0.04 & 0.09 & 0.05 \\
      \midrule
      \rowcolor{gray!10}
        & LLMEval(A.)   & 0.48 & 0.19 & 0.49 & 0.50 & 0.27 
                        & 0.12 & 0.21 
                        & 0.13 & 0.04 & 0.10 \\
      \rowcolor{gray!10}
        & LLMEval(I.)   & 0.46 & 0.20 & 0.50 & 0.50 & 0.26 
                        & 0.10 & 0.18 
                        & 0.19 & 0.14 & 0.18 \\
      \rowcolor{gray!10}
      \multirow{-3}{*}{Eli5}
        & BERTScore     & 0.58 & 0.27 & 0.57 & 0.59 & -0.01 
                        & -0.03 & 0.08 
                        & 0.04 & 0.06 & 0.07 \\
      \midrule
      \multirow{-1}{*}{Fact-Bio}
        & FActScore     & 0.78 & 0.31 & 0.78 & 0.75 & 0.30 
                        & 0.65 & 0.21 
                        & 0.45 & 0.56 & 0.54 \\
      \midrule
      \rowcolor{gray!10}
        & LLMEval(A.)   & 0.42 & 0.17 & 0.43 & 0.45 & 0.23 
                        & 0.23 & 0.13 
                        & 0.10 & 0.08 & 0.13 \\
      \rowcolor{gray!10}
        & LLMEval(I.)   & 0.42 & 0.16 & 0.39 & 0.43 & 0.13 
                        & 0.24 & 0.16 
                        & 0.20 & 0.22 & 0.24 \\
      \rowcolor{gray!10}
      \multirow{-3}{*}{MED LF QA}
        & BERTScore     & 0.55 & 0.38 & 0.52 & 0.54 & 0.16 
                        & 0.22 & 0.36 
                        & 0.09 & 0.04 & 0.05 \\
      \midrule
      \multirow{2}{*}{Opus-100}
        & GEMBA         & 0.50 & 0.36 & 0.44 & 0.49 & 0.28 
                        & 0.33 & 0.21 
                        & 0.16 & 0.21 & 0.22 \\
      \multirow{-2}{*}{Opus-100}
        & BERTScore     & 0.66 & 0.50 & 0.63 & 0.65 & 0.23 
                        & 0.11 & 0.15 
                        & 0.06 & 0.05 & -0.02 \\
      \midrule
      \rowcolor{gray!10}
        & G-Eval(Coh.)  & 0.01 & 0.05 & 0.03 & 0.02 & 0.16 
                        & 0.05 & 0.04 
                        & 0.06 & -0.01 & -0.02 \\
      \rowcolor{gray!10}
        & G-Eval(Con.)  & 0.01 & 0.09 & 0.10 & 0.05 & 0.15 
                        & 0.08 & 0.02 
                        & 0.02 & 0.03 & -0.01 \\
      \rowcolor{gray!10}
        & G-Eval(Flu.)  & 0.09 & 0.01 & 0.10 & 0.05 & 0.03 
                        & 0.03 & 0.03 
                        & 0.00 & -0.02 & 0.02 \\
      \rowcolor{gray!10}
        & G-Eval(Rel.)  & 0.15 & 0.09 & 0.17 & 0.13 & 0.14 
                        & 0.18 & 0.05 
                        & 0.04 & 0.01 & 0.06 \\
      \rowcolor{gray!10}
      \multirow{-5}{*}{Summeval}
        & BERTScore     & 0.56 & 0.24 & 0.57 & 0.52 & 0.08 
                        & 0.16 & 0.00 
                        & -0.03 & 0.04 & -0.02 \\
      \midrule
        & LLMEval(A.)   & 0.53 & 0.27 & 0.55 & 0.55 & 0.08 
                      & 0.25 & 0.22 
                      & 0.04 & 0.08 & 0.06 \\
      & LLMEval(I.)   & 0.48 & 0.24 & 0.45 & 0.49 & 0.06 
                      & 0.14 & 0.15 
                      & 0.07 & 0.04 & 0.12 \\
    \multirow{-3}{*}{TruthfulQA}
      & BERTScore     & 0.60 & 0.13 & 0.39 & 0.46 & 0.08 
                      & 0.12 & 0.11 
                      & 0.03 & 0.10 & -0.02 \\
      \midrule
      \multicolumn{2}{|l|}{AVERAGE}
        & 0.43 & 0.25 & 0.41 & 0.42 & 0.17 
        & 0.17 & \textbf{0.13} 
        & 0.09 & 0.10 & 0.08 \\
      \bottomrule
    \end{tabular}
  }
  \caption{\textbf{Pearson Correlation Benchmark.} We compare two baselines: 1) confidence-based regression (CE-Reg) uses 7 CE features (five consistency-based, two verbalized); and 2) reference-free LLM-as-a-Judge (RF-LLMaaJ), evaluated in 4-, 8-, and 16-shot settings. For RMSE, lower is better. Judge(Func.) = CodeJudge Functional Correctness. Judge(Incon.) = CodeJudge Inconsistency. LLMEval(A.) = LLM Eval Accuracy. LLMEval(I.) = LLMEval Informativeness.}
  \label{tab:benchmark_corr}
\end{table*}

%% file: table/ac_bc_table_ace.tex
\begin{table*}[!t]
  \centering
  \makebox[\textwidth][l]{%
    \hspace*{-0.5cm}
  \scriptsize
  \setlength{\tabcolsep}{4pt}
  \renewcommand{\arraystretch}{1.2}
  
    \begin{tabular}{|
        l|l|
        *{5}{c}|                
        *{2}{>{\centering\arraybackslash}p{1.2cm}}|  
        *{3}{>{\centering\arraybackslash}p{1.0cm}}|  
      }
      \midrule
      \multicolumn{2}{|c|}{} 
        & \multicolumn{5}{c|}{\textbf{CE$-$Reg: Consistency-based}}
        & \multicolumn{2}{c|}{\textbf{CE$-$Reg: Verbalized}}
        & \multicolumn{3}{c|}{\textbf{RF-LLMaaJ}} \\
      \cline{3-12}
    \textbf{Dataset}  & \textbf{Metric}  
      & \textbf{DegMat.} & \textbf{Eccen.} & \textbf{EigVal.}  
      & \textbf{LexSim.} & \textbf{NumSet.}  
      & \textbf{1S}   & \textbf{2S}  
      & \textbf{4-shot} & \textbf{8-shots} & \textbf{16-shot} \\
    \midrule
    \multirow{3}{*}{APIGen}
      & Judge(Func.)  & -0.09 & -0.53 & -0.49 & -0.36 & -0.95 
                      & -0.94 & -0.93 
                      & -0.36 & -0.34 & -0.34 \\
      & Judge(Incon.) & -0.14 & -0.56 & -0.52 & -0.42 & -0.89 
                      & -0.94 & -0.86 
                      & -0.59 & -0.60 & -0.56 \\
      & CodeBLEU      & -0.09 & -0.54 & -0.48 & -0.36 & -0.95 
                      & -0.94 & -0.92 
                      & -0.37 & -0.35 & -0.36 \\
    \midrule
    \multirow{3}{*}{ASQA}
      & LLMEval(A.)   & -0.18 & -0.66 & -0.15 & -0.16 & -0.95 
                      & -0.87 & -0.82 
                      & -0.37 & -0.39 & -0.38 \\
      & LLMEval(I.)   & -0.21 & -0.68 & -0.24 & -0.23 & -0.95 
                      & -0.89 & -0.81 
                      & -0.41 & -0.40 & -0.46 \\
      & BERTScore     & -0.24 & -0.68 & -0.22 & -0.27 & -0.90 
                      & -0.86 & -0.79 
                      & -0.34 & -0.33 & -0.38 \\
    \midrule
    \multirow{1}{*}{BiGGen}
      & Judge         & -0.15 & -0.77 & -0.26 & -0.23 & -0.89 
                      & -0.80 & -0.68 
                      & -0.52 & -0.51 & -0.51 \\
    \midrule
    \multirow{3}{*}{BigCodeBench}
      & Judge(Func.)  & -0.20 & -0.37 & -0.20 & -0.18 & -0.92 
                      & -0.94 & -0.78 
                      & -0.39 & -0.42 & -0.40 \\
      & Judge(Incon.) & -0.14 & -0.45 & -0.17 & -0.13 & -0.95 
                      & -0.93 & -0.92 
                      & -0.49 & -0.52 & -0.49 \\
      & CodeBLEU      & -0.19 & -0.48 & -0.23 & -0.22 & -0.91 
                      & -0.90 & -0.76 
                      & -0.43 & -0.43 & -0.38 \\
    \midrule
    \multirow{3}{*}{Eli5}
      & LLMEval(A.)   & -0.20 & -0.91 & -0.22 & -0.21 & -0.87 
                      & -0.92 & -0.83 
                      & -0.36 & -0.35 & -0.33 \\
      & LLMEval(I.)   & -0.18 & -0.90 & -0.19 & -0.17 & -0.95 
                      & -0.93 & -0.76 
                      & -0.22 & -0.22 & -0.18 \\
      & BERTScore     & -0.22 & -0.86 & -0.22 & -0.22 & -0.88 
                      & -0.85 & -0.76 
                      & -0.63 & -0.57 & -0.55 \\
    \midrule
    \multirow{1}{*}{Fact-Bio}
      & FActScore     & -0.14 & -0.86 & -0.13 & -0.11 & -0.90 
                      & -0.82 & -0.81 
                      & -0.40 & -0.39 & -0.38 \\
    \midrule
    \multirow{3}{*}{MED LF QA}
      & LLMEval(A.)   & -0.18 & -0.84 & -0.19 & -0.19 & -0.95 
                      & -0.85 & -0.84 
                      & -0.37 & -0.34 & -0.33 \\
      & LLMEval(I.)   & -0.17 & -0.84 & -0.19 & -0.18 & -0.59 
                      & -0.95 & -0.65 
                      & -0.29 & -0.30 & -0.29 \\
      & BERTScore     & -0.19 & -0.83 & -0.22 & -0.22 & -0.93 
                      & -0.90 & -0.85 
                      & -0.49 & -0.45 & -0.45 \\
    \midrule
    \multirow{2}{*}{Opus-100}
      & GEMBA         & -0.17 & -0.45 & -0.30 & -0.24 & -0.94 
                      & -0.84 & -0.80 
                      & -0.44 & -0.40 & -0.38 \\
      & BERTScore     & -0.10 & -0.45 & -0.25 & -0.21 & -0.91 
                      & -0.89 & -0.80 
                      & -0.40 & -0.34 & -0.32 \\
    \midrule
    \multirow{5}{*}{Summeval}
      & G-Eval(Coh.)  & -0.18 & -0.88 & -0.18 & -0.20 & -0.95 
                      & -0.94 & -0.92 
                      & -0.78 & -0.81 & -0.86 \\
      & G-Eval(Con.)  & -0.06 & -0.82 & -0.07 & -0.07 & -0.95 
                      & -0.95 & -0.87 
                      & -0.44 & -0.40 & -0.40 \\
      & G-Eval(Flu.)  &  0.04 & -0.82 &  0.04 &  0.04 & -0.95 
                      & -0.92 & -0.10 
                      & -0.29 & -0.26 & -0.24 \\
      & G-Eval(Rel.)  & -0.17 & -0.87 & -0.19 & -0.19 & -0.95 
                      & -0.94 & -0.93 
                      & -0.61 & -0.58 & -0.57 \\
      & BERTScore     & -0.20 & -0.86 & -0.20 & -0.21 & -0.90 
                      & -0.89 & -0.81 
                      & -0.51 & -0.52 & -0.50 \\
    \midrule
    \multirow{3}{*}{TruthfulQA}
      & LLMEval(A.)   & -0.16 & -0.70 & -0.25 & -0.20 & -0.88 
                      & -0.92 & -0.79 
                      & -0.38 & -0.35 & -0.37 \\
      & LLMEval(I.)   & -0.23 & -0.55 & -0.29 & -0.28 & -0.63 
                      & -0.76 & -0.74 
                      & -0.44 & -0.43 & -0.42 \\
      & BERTScore     & -0.19 & -0.80 & -0.28 & -0.25 & -0.86 
                      & -0.81 & -0.71 
                      & -0.52 & -0.50 & -0.50 \\
    \midrule
    \multicolumn{2}{|l|}{AVERAGE}
      & \textbf{-0.16} & -0.70 & -0.23 & -0.21 & -0.90 
      & -0.89 & -0.79 
      & -0.44 & -0.43 & -0.42 \\
    \bottomrule
\end{tabular}

  }
  \caption{\textbf{ACE(95\%) Benchmark.} We compare two baselines: 1) confidence-based regression (CE-Reg) uses 7 CE features (five consistency-based, two verbalized); and 2) reference-free LLM-as-a-Judge (RF-LLMaaJ), evaluated in 4-, 8-, and 16- settings. For ACE(95 \%), values closer to zero are better. Negative values indicate intervals that are too narrow (over-confidence), while positive values indicate intervals that are too wide (under-confidence).
  Judge(Func.) = CodeJudge Functional Correctness. Judge(Incon.) = CodeJudge Inconsistency. LLMEval(A.) = LLM Eval Accuracy. LLMEval(I.)  = LLMEval Informativeness. 
  }
  \label{tab:benchmark_ace}
\end{table*}